\documentclass{article}
\usepackage{microtype}
\usepackage{booktabs}
\usepackage[nohyperref,preprint]{icml2026}

\usepackage{algorithm}
\usepackage{algorithmicx}
\usepackage[noend]{algpseudocode}
\usepackage{amsmath}
\usepackage{amssymb}
\usepackage{amsthm}
\usepackage{bbm}
\usepackage{bm}
\usepackage{caption}
\usepackage{color}
\usepackage{colortbl}
\usepackage{dirtytalk}
\usepackage{dsfont}
\usepackage{enumerate}
\usepackage{graphicx}
\usepackage{listings}
\usepackage{mathtools}
\usepackage{natbib}
\usepackage{soul}
\usepackage{subcaption}
\usepackage{tcolorbox}
\usepackage{times}
\usepackage{wrapfig}
\usepackage{xspace}

\usepackage[dvipsnames]{xcolor}
\usepackage[colorlinks=true]{hyperref}
\usepackage[capitalize,noabbrev]{cleveref}

\definecolor{darkblue}{rgb}{0.0,0.0,0.65}
\definecolor{darkred}{rgb}{0.65,0.0,0.0}
\definecolor{darkgreen}{rgb}{0.0,0.5,0.0}
\definecolor{tab:blue}{RGB}{31,119,180}  %
\definecolor{tab:red}{RGB}{214,39,40}  %
\definecolor{tab:green}{RGB}{44,160,44}  %
\definecolor{tab:orange}{RGB}{255,127,14}  %
\definecolor{highlight}{gray}{0.92} %
\hypersetup{
  linktocpage=true,
  colorlinks=true,
  citecolor=darkblue,
  filecolor=darkblue,
  linkcolor=darkred,
  urlcolor=darkblue
}

\usepackage{tikz-cd}
\usetikzlibrary{positioning,arrows.meta}

\usepackage{amsmath,amsfonts,bm,bbm,mathrsfs,dsfont}
\usepackage{amssymb,mathtools}

\newcommand{\indicator}{\mathds{1}}

\DeclareMathAlphabet{\mathsfit}{\encodingdefault}{\sfdefault}{m}{sl}
\SetMathAlphabet{\mathsfit}{bold}{\encodingdefault}{\sfdefault}{bx}{n}

\def\gD{{\mathcal{D}}}

\def\gJ{{\mathcal{J}}}

\def\sN{{\mathbb{N}}}

\def\sP{{\mathbb{P}}}

\newcommand{\E}{\mathbb{E}}

\DeclareMathOperator*{\argmax}{arg\,max}

\def\1{\mathbbm{1}}

\usepackage{amsthm}
\theoremstyle{plain}

\newtheorem{lemma}{Lemma}

\usepackage[textsize=tiny]{todonotes}

\mathchardef\mhyphen="2D
\newcommand{\fem}{\ensuremath{\color{Green}\tt FEM}\xspace}
\newcommand{\algrs}{\ensuremath{\color{Green}\tt RS}\xspace}
\newcommand{\algrsone}{\ensuremath{\color{Green}\tt RS\mhyphen 1}\xspace}
\newcommand{\algrsthree}{\ensuremath{\color{Green}\tt RS\mhyphen 3}\xspace}
\newcommand{\algrsfive}{\ensuremath{\color{Green}\tt RS\mhyphen 5}\xspace}
\newcommand{\algstar}{\ensuremath{\color{Green}\tt STaR}\xspace}
\newcommand{\algstartwo}{\ensuremath{\color{Green}\tt STaR2}\xspace}
\newcommand{\algstarthree}{\ensuremath{\color{Green}\tt STaR3}\xspace}
\newcommand{\algpps}{\ensuremath{\color{Green}\tt PPS}\xspace}
\newcommand{\algppstwo}{\ensuremath{\color{Green}\tt PPS2}\xspace}
\newcommand{\algppsthree}{\ensuremath{\color{Green}\tt PPS3}\xspace}

\icmltitlerunning{Learning to Reason in LLMs by Expectation Maximization}

\begin{document}

\twocolumn[
\icmltitle{Learning to Reason in LLMs by Expectation Maximization}

\icmlsetsymbol{equal}{*}

\begin{icmlauthorlist}
\icmlauthor{Junghyun Lee}{kaist-ai}
\icmlauthor{Branislav Kveton}{adobe}
\icmlauthor{Anup Rao}{adobe}
\icmlauthor{Subhojyoti Mukherjee}{adobe}
\icmlauthor{Ryan A. Rossi}{adobe}
\icmlauthor{Sunav Choudhary}{adobe}
\icmlauthor{Alexa Siu}{adobe}
\end{icmlauthorlist}

\icmlaffiliation{kaist-ai}{Kim Jaechul Graduate School of AI, KAIST, Seoul, Republic of Korea}
\icmlaffiliation{adobe}{Adobe Research, San Jose, CA, USA}

\icmlcorrespondingauthor{Junghyun Lee}{jh\_lee00@kaist.ac.kr}

\vskip 0.3in
]

\printAffiliationsAndNotice{}

\begin{abstract}
Large language models (LLMs) solve reasoning problems by first generating a rationale and then answering. We formalize reasoning as a latent variable model and derive a reward-based filtered expectation-maximization (FEM) objective for learning to reason. This view connects EM and modern reward-based optimization, and shows that the main challenge lies in designing a sampling distribution of rationales that justify correct answers. We instantiate and compare three sampling schemes: rejection sampling with a budget, self-taught reasoner (STaR), and prompt posterior sampling (PPS), which only keeps the rationalization stage of STaR that conditions on the correct answer in the prompt. We experiment with LLM-as-a-judge calibration and summarization from feedback tasks, where conditioning on the correct answer provides a strong guidance for generating rationales. Our experiments show the efficacy of PPS over other sampling schemes, and that the sampling scheme can have a significant impact on performance.
\end{abstract}

\section{Introduction}
\label{sec:introduction}

\emph{Large language models (LLMs)} have shown strong performance on reasoning tasks, such as math word problems and multiple-choice exams~\citep{zelikman2022STaR,shao24deepseekmath}. In these tasks, it is often difficult to answer questions directly. Instead, it is more beneficial to first generate an intermediate \emph{reasoning trace} or \emph{rationale}, and then output the answer. This is the key idea in \emph{chain-of-thought (CoT)} prompting, where the model is asked to \emph{think step by step} before answering~\citep{CoT,CoT2}.

\textbf{Reasoning as a latent variable model (LVM).} We consider the task of learning a mapping from a question $x$ to its answer $y^\star$, and specifically focus on deterministically verifiable answers. This setting is common in domains like math reasoning~\citep{deepseek-r1}, multiple-choice question-answering~\citep{mmlu}, and learning LLM judges or reward models from human feedback~\citep{chiang25tract,ankner24critiqueoutloud,sahoo2025llmjudge}. When the task is complex, it is hard to model $x \rightarrow y^\star$ directly. This motivates the introduction of a rationale -- a \emph{latent variable} $z$ representing the (unobserved) intermediate reasoning process. In words, conditioned on both the question $x$ and rationale $z$, the model has a much higher chance of generating the correct answer $y^\star$ than when conditioned on $x$ alone. With the rationale, the predictive structure factors as a \emph{latent variable model (LVM)}~\citep{koller2009PGM}:
\begin{equation} \label{eq:markov chain}
    \begin{tikzcd}[column sep=1.5em, baseline=(current bounding box.center)]
        x \arrow[r] \arrow[rr, bend left=25] & z \arrow[r] & y^\star
    \end{tikzcd}
\end{equation}
This latent-variable perspective on rationales has been explored in various forms in prior works~\citep{zelikman2022STaR,yuan2023RFT,phan2023TRICE,hu2024amortizing,zhong2025BRiTE,tang2025JEPO,xu2025EM}. We adopt it as a starting point for understanding existing \emph{learning to reason} algorithms and, in particular, the role of sampling.

\textbf{Expectation maximization (EM).} Once rationales are modeled as latent variables, the \emph{EM algorithm}~\citep{dempster1977EM,moon1996EM,neal1998EM} can be used to learn them. The algorithm alternates between computing a posterior distribution over rationales conditioned on the observed question-answer pairs (\textbf{E-step}) and maximizing the expected complete data log-likelihood (\textbf{M-step}). Neither of the steps can be implemented exactly in LLMs. The \textbf{E-step} requires computing an expectation over $z$ conditioned on both $x$ and $y^\star$ in \eqref{eq:markov chain}. The expectation does not have a closed form, and approximating it via Monte Carlo sampling requires sampling from the exact posterior. In theory, classic statistical techniques, such as rejection sampling~\citep{neal1998EM} and Markov chain Monte Carlo~\citep{MCEM,doucet01sequential,phan2023TRICE}, could be used, but they are not guaranteed to be computationally or statistically inefficient. The \textbf{M-step} also lacks a closed-form solution and is typically approximated by gradient-based optimization~\citep{bottou2018survey}. Many prior works~\citep{balakrishnan2017EM,neath2013convergence,cappe09online} studied this and similar approaches.

\textbf{Self-improvement in LLMs.} Concurrently, the LLM community has introduced algorithms that train LLMs by explicitly generating and filtering their own reasoning traces, a paradigm known as \emph{self-improvement}. Examples include rejection-based fine-tuning~\citep{yuan2023RFT,dong2023RAFT,singh2024ReSTEM,shao24deepseekmath} and self-taught reasoning~\citep{zelikman2022STaR}. These methods share a similar high-level structure:
\begin{enumerate}
  \item Sample one or more rationale-answer pairs $z, \hat{y}$ conditioned on the question $x$;
  \item Verify if $\hat{y} = y^\star$;
  \item Fine-tune on accepted rationale-answer pairs.
\end{enumerate}
From the LVM perspective, these methods perform approximate \textbf{E-steps} followed by \textbf{M-steps}. While \citet{zelikman2022STaR} noted this link, they focused on a policy-gradient justification for their proposed self-taught reasoner, leaving the formal EM connection under-explored.

\textbf{Contributions.} In this work, we examine this connection in detail. Our contributions are as follows:
\begin{itemize}
  \item \textbf{Formalizing reasoning as filtered EM.} We revisit LLM reasoning as an LVM and derive a reward-based \emph{filtered EM} (\fem) objective. We approximate the \textbf{E-step} via a single Monte Carlo sample and the \textbf{M-step} using a \emph{filtered} gradient update. At a high level, we sample rationale-answer pairs from the latest model for all questions, keep them if the sampled answer is correct, and then fine-tune on the retained pairs. This formulation rigorously relates EM to modern reward-based optimization and frames LLM self-improvement as an iterative supervised fine-tuning procedure where latent rationales are resampled at each epoch. We also highlight its connection to reward-weighted fine-tuning of~\citet{mukherjee2025multiturn}.
  \item \textbf{Unifying and instantiating sampling schemes.} We demonstrate that existing \emph{learning to reason} algorithms, such as rejection sampling (\algrs) and self-taught reasoning (\algstar), can be viewed as specific instances of our \fem framework with different rationale sampling schemes. Furthermore, we propose prompt posterior sampling (\algpps), which can be viewed as the second rationalization stage of \algstar and, to the best of our knowledge, has not yet been systematically studied.
  \item \textbf{Empirical evaluation of sampling schemes.} We empirically evaluate all sampling schemes on LLM-as-a-judge calibration~\citep{zheng23judging,sahoo2025llmjudge} and summarization from feedback~\citep{stiennon2020summarize} tasks. The evaluation is conducted on two Llama~\citep{Llama} and three Qwen~\citep{Qwen2.5} models. Unlike in complex math reasoning~\citep{shao24deepseekmath}, conditioning on the correct answer provides a strong guiding signal for rationale generation in our tasks, and this is why we experiment with them. We observe that \algpps consistently outperforms other sampling schemes. Our results show that the \emph{sampling scheme} approximating the \textbf{E-step} is critical to the effectiveness of self-improvement in LLMs.
\end{itemize}

\section{EM and Learning to Reason}
\label{sec:em}

\begin{algorithm*}[t!]
  \caption{Filtered EM (\fem) for learning to reason.}
  \label{alg:EM-filter}
  \begin{algorithmic}[1]
    \State \textbf{Input:} Number of iterations $K$, dataset $\gD = \{(x_i, y_i^\star)\}_{i = 1}^N$, initial model $\theta^{(0)}$, sampling distribution $q(\cdot \mid x, y^\star; \theta)$
    \For{$k = 1, \dots, K$}
      \ForAll{$i \in [N]$}
        $\hat{z}_i^{(k)}, \hat{y}_i^{(k)}
        \sim q(\cdot \mid x_i, y_i^\star; \theta^{(k - 1)})$
        \Comment{Rationale sampling}
      \EndFor
      \State $\theta^{(k)} \gets \theta^{(k - 1)}$
      \ForAll{$i \in [N]$}
        $\theta^{(k)}
        \gets \theta^{(k)} + \eta^{(k)} r(\hat{y}_i^{(k)}, y_i^\star)
        \nabla \log \pi(\hat{z}_i^{(k)}, \hat{y}_i^{(k)} \mid x_i; \theta^{(k)})$
        \Comment{Gradient update}
      \EndFor
    \EndFor
    \State \textbf{Output:} Learned model $\theta^{(K)}$
  \end{algorithmic}
\end{algorithm*}

Let $\gD = \{(x_i, y_i^\star)\}_{i = 1}^N$ be a dataset of $N$ question-answer pairs. We denote the probability that an LLM $\pi(\cdot \mid \cdot; \theta)$ with parameters $\theta$ generates an answer $y$ when prompted with $x$ by $\pi(y \mid x; \theta)$. For reasoning models that generate rationales, we write $\pi(z, y \mid x; \theta)$ for the joint probability distribution over a rationale $z$ and final answer $y$. We denote by $\nabla$ the gradient operator with respect to $\theta$.

As discussed in detail in \cref{sec:introduction}, we model reasoning using a latent variable $z_i$ because learning $x_i \rightarrow z_i, y_i^\star$ leads to more accurate answers than $x_i \rightarrow y_i^\star$ alone. Intuitively, $z_i$ is the underlying reasoning, and therefore, conditioned on both $x_i$ and $z_i$, the LLM is more likely to generate correct answers $y_i^\star$. A standard way of learning LVMs is the \emph{EM algorithm}~\citep{dempster1977EM}. EM is an iterative algorithm with two steps in each iteration. The \textbf{E-step} computes the expected complete data log-likelihood given the most recent model parameter $\theta^{(k - 1)}$,
\begin{equation*}
  \textstyle
  \gJ(\theta)
  := \sum_{i = 1}^N \E_{z_i \sim \pi(\cdot \mid x_i, y_i^\star; \theta^{(k - 1)})}[\log \pi(z_i, y_i^\star \mid x_i; \theta)]\,,
\end{equation*}
where $\pi(\cdot \mid x_i, y_i^\star; \theta^{(k - 1)})$ is the posterior distribution of the latent rationale $z_i$ under the previous iterate $\theta^{(k - 1)}$. The \textbf{M-step} solves for $\theta^{(k)} = \argmax_\theta \gJ(\theta)$.

\subsection{From EM to Reward-Weighted Fine-Tuning}
\label{sec:em to fine-tuning}

We first focus on the \textbf{M-step} and introduce a series of approximations that lead our algorithm. First, we approximate the exact maximization by a single step of gradient ascent
\begin{equation}
  \theta^{(k)}
  \gets \theta^{(k)} + \eta^{(k)} \underbrace{\nabla \,
  \E_{\hat{z}_i}[\log \pi(\hat{z}_i, y_i^\star \mid x_i; \theta)]}_{(*)}
  \label{eqn:EM-pop}
\end{equation}
for each data point $i \in [N]$, where $\eta^{(k)}$ is the learning rate at iteration $k$ and the expectation $\E_{\hat{z}_i}$ is taken with respect to $\hat{z}_i \sim \pi(\cdot \mid x_i, y_i^\star; \theta^{(k - 1)})$.

Now note that the posterior distribution is by definition \emph{consistent} with the evidence that we condition on. Therefore, any $\hat{z}_i, \hat{y}_i$ drawn from $\pi(\cdot \mid x_i, y_i^\star)$ must satisfy $\hat{y}_i = y_i^\star$ almost surely. This implies that checking whether the generated answer matches the ground truth is theoretically equivalent to enforcing the posterior distribution's support. Thus, even if one were to sample from a broader proposal distribution, applying a consistency filter ensures that the update operates only on answer-consistent samples. We formalize this using a binary reward $r(\hat{y}_i, y_i^\star) := \indicator[\hat{y}_i = y_i^\star]$, which lets us rewrite \eqref{eqn:EM-pop} with
\begin{equation}
  (*)
  = \nabla \, \E_{\hat{z}_i, \hat{y}_i}[r(\hat{y}_i, y_i^\star)
  \log \pi(\hat{z}_i, \hat{y}_i \mid x_i; \theta)]\,,
\end{equation}
where the expectation $\E_{\hat{z}_i, \hat{y}_i}$ is taken with respect to $\hat{z}_i, \hat{y}_i \sim \pi(\cdot \mid x_i, y_i^\star; \theta^{(k - 1)})$.

We make two additional observations. First, the expectation is defined with respect to a distribution that depends only on the previous parameters $\theta^{(k - 1)}$, but not the optimized $\theta$. Second, the reward does not depend on $\theta$. As a result, the gradient can be moved inside the expectation and we get
\begin{equation}
  (*)
  = \E_{\hat{z}_i, \hat{y}_i}[r(\hat{y}_i, y_i^\star)
  \nabla \log \pi(\hat{z}_i, \hat{y}_i \mid x_i; \theta)]\,.
  \label{eqn:EM-reward-2}
\end{equation}
In LLMs, this expectation cannot be computed efficiently because $\hat{z}_i, \hat{y}_i$ are sequences of tokens from an exponentially large space. Therefore, we approximate it by a single Monte Carlo sample as
\begin{equation}
  (*)
  \approx r(\hat{y}_i, y_i^\star)
  \nabla \log \pi(\hat{z}_i, \hat{y}_i \mid x_i; \theta)\,,
  \label{eqn:EM-filter}
\end{equation}
where $\hat{z}_i, \hat{y}_i \sim \pi(\cdot \mid x_i, y_i^\star; \theta^{(k - 1)})$ is a single rationale-answer sample. This is a \emph{filtered} gradient update of the classic Monte-Carlo EM~\citep{MCEM,neath2013convergence}, where the reward $r(\hat{y}_i, y_i^\star)$ serves as a filter in broader proposal distributions (\cref{sec:sampling}).

\begin{figure*}[t]
  \centering
  \begin{minipage}[t]{0.48\textwidth}
\begin{algorithm}[H]
  \caption{\algrs}
  \label{alg:rejection-sampling}
  \begin{algorithmic}[1]
    \State \textbf{Input:} Data point $(x, y^\star)$, budget $M \in \sN$
    \For{$m = 1, \dots, M$}
       \State $\hat{z}_m, \hat{y}_m \sim \pi(\cdot \mid x; \theta)$
       \If{$\hat{y}_m = y^\star$}
         \State \textbf{Output:} $\hat{z}_m, y^\star$
       \EndIf
    \EndFor
    \State \textbf{Output:} $\hat{z}_M, \hat{y}_M$
  \end{algorithmic}
\end{algorithm}
  \end{minipage}
  \hfill
  \begin{minipage}[t]{0.48\textwidth}
\begin{algorithm}[H]
  \caption{\algstar}
  \label{alg:star-sampling}
  \begin{algorithmic}[1]
    \State \textbf{Input:} Data point $(x, y^\star)$
    \State $\hat{z}, \hat{y} \gets {\algrs}((x, y^\star), 1)$
    \Comment{Rejection sampling}
    \If{$\hat{y} = y^\star$}
      \State \textbf{Output:} $\hat{z}, \hat{y}$
    \Else
      \State $\hat{z}', \hat{y}' \sim q_{\textsc{pps}}(\cdot \mid x, y^\star; \theta)$
      \Comment{Rationalization}
    \EndIf
    \State \textbf{Output:} $\hat{z}', \hat{y}'$
  \end{algorithmic}
\end{algorithm}
  \end{minipage}
\end{figure*}

\subsection{Practical Meta-Algorithm}
\label{sec:algorithm}

Exact sampling of $\hat{z}_i, \hat{y}_i$ from the posterior distribution in \eqref{eqn:EM-filter} is computationally infeasible in LLMs. This is because sampling from the posterior requires inverting the generative process in \eqref{eq:markov chain} to obtain reasoning paths $\hat{z}_i$ leading to correct answers $y_i^\star$. Moreover, $\hat{z}_i, \hat{y}_i$ are sequences of tokens from an exponentially large space.

We address this challenge by considering a generic \emph{rationale proposal distribution} $q(\hat{z}, \hat{y} \mid x, y^\star; \theta)$ and implement it by \emph{prompting} the most recent model $\theta^{(k - 1)}$. Then we approximate \eqref{eqn:EM-filter} by sampling $\hat{z}_i, \hat{y}_i \sim q(\cdot \mid x_i, y_i^\star; \theta^{(k - 1)})$. More specifically, for each data point, we sample $\hat{z}_i, \hat{y}_i$; keep it only if the sampled answer is correct, $\hat{y}_i = y_i^\star$; and then fine-tune on retained $\hat{z}_i, \hat{y}_i$. Because the reward $r(\hat{y}_i, y_i^\star)$ is binary, the rationales leading to incorrect answers are filtered out in the gradient update.

The pseudo-code of our algorithm is given in \cref{alg:EM-filter}. We call it \emph{filtered EM (\fem)} due to reward filtering in \eqref{eqn:EM-filter}. The algorithm is iterative. In each iteration, it applies the E- and M-steps (\cref{sec:em to fine-tuning}) to all question-answer pairs sequentially: Line 3 is the \textbf{E-step} applied to dataset $\gD$ and Line 5 is the corresponding \textbf{M-step} with reward filtering. For simplicity of exposition, we assume a batch size of $1$. The design of \fem is motivated by \emph{supervised fine-tuning (SFT)}. In particular, each iteration of \fem can be viewed as an epoch in SFT, where the training data are adapted to the learned model due to sampling $\hat{z}_i^{(k)}, \hat{y}_i^{(k)}$ in Line 3. It can also be viewed as reward-weighted fine-tuning \citep{mukherjee2025multiturn}, where the model is updated to increase the reward-weighted log-likelihood of the sampled rationale-answer pairs.

The rationale proposal distribution $q(\cdot \mid x, y^\star; \theta)$ is a design decision and should have two properties. First, the probability of event $\hat{y} = y^\star$ should be maximized to maximize the amount of training data. Second, the quality of rationales $\hat{z}$ should be high to ensure that conditioning on $x, \hat{z}$ leads to better predicted answers $\hat{y}$ and generalization. We provide different instantiations of $q$ in \cref{sec:sampling}.

\subsection{Filtered EM and Reinforcement Learning}
\label{sec:rl}

The filtered EM update can also be justified through its relationship to reward-based maximization in \emph{reinforcement learning (RL)}. To show this, we build on the work of \citet{mukherjee2025multiturn}, which extends results of \citet{ma19imitationregularized,liang2022local} from offline logged bandits to offline RL.

\begin{lemma}
\label{lem:mukherjee} Let $x, y^\star$ be a question-answer pair and $r(\hat{y}, y^\star) = \indicator[\hat{y} = y^\star]$. Then for any parameter vector $\theta$ and iteration $k \geq 1$,
\begin{align*}
  & \underbrace{\E_{\hat{z}, \hat{y} \sim \pi(\cdot \mid x; \theta)}\!\left[ r(\hat{y}, y^\star) \right]}_{\text{RHS}} \\
  & \geq \underbrace{\E_{\hat{z}, \hat{y} \sim q(\cdot \mid x, y^\star; \theta^{(k - 1)})}\!\left[ r(\hat{y}, y^\star) \log \pi(\hat{z}, \hat{y} \mid x; \theta) \right]}_{\text{LHS}} {} + C\,.
\end{align*}
where $C \geq 0$ is independent of $\theta$. The inequality is tight when $q(\cdot \mid x, y^\star; \theta^{(k - 1)}) \equiv \pi(\cdot \mid x; \theta)$.
\end{lemma}
\begin{proof}
The proof mimics~\citet{mukherjee2025multiturn} and is provided in \cref{sec:proof} for completeness.
\end{proof}

The above lemma says that reward-weighted fine-tuning (LHS) essentially optimizes a lower bound on the reward maximization objective (RHS). Our \fem update in \eqref{eqn:EM-filter} iteratively refines it by alternating between updating the data sampling distribution $q$ (\textbf{E-step}) and maximizing the lower bound (\textbf{M-step}).

\section{Rationale Sampling}
\label{sec:sampling}

Now we instantiate the rationale proposal distribution $q(\cdot \mid x, y^\star; \theta)$ in \cref{alg:EM-filter}. We focus on three representative schemes: rejection sampling with a budget, self-taught reasoning, and prompt posterior sampling.

\textbf{Rejection sampling with budget $M$ (\algrs).} A statistically correct way of sampling $z$ from the LVM in \eqref{eq:markov chain} is rejection sampling~\citep{neal1998EM}. In LLMs, it corresponds to sampling reasoning-answer pairs conditioned on the question until the correct answer is sampled, with a budget of at most $M$. The underlying intuition is to create a virtuous cycle: as the model improves, it is more likely to sample correct answers, thereby providing more training data for the next \textbf{M-step}. This approach has been proposed before under the names of rejection fine-tuning~\citep{yuan2023RFT,dong2023RAFT,singh2024ReSTEM} and online rejection sampling fine-tuning~\citep[Appendix A.1.3]{shao24deepseekmath}. The pseudo-code of this method is given in \cref{alg:rejection-sampling}.

\begin{figure}[t!]
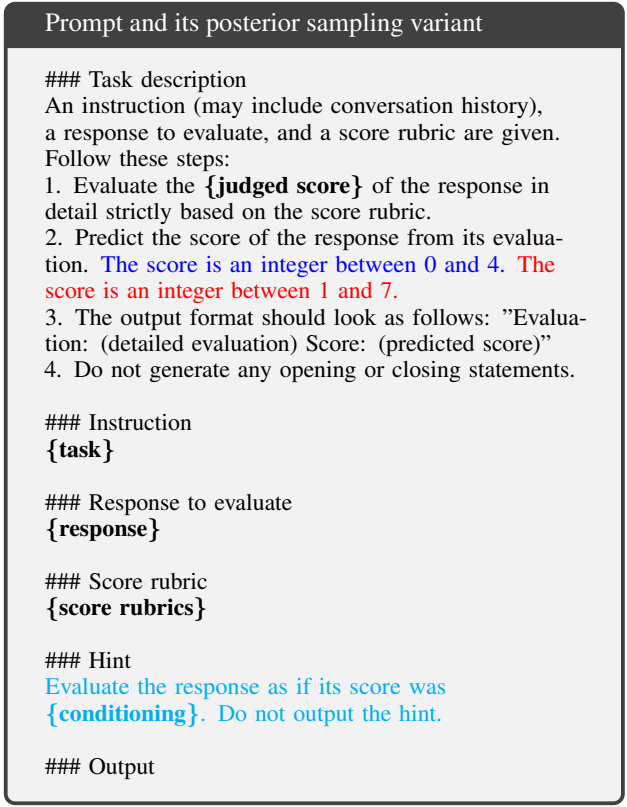

  \centering
  \begin{tcolorbox}[title=Prompt and its posterior sampling variant,halign=left,fontupper=\small,fontlower=\small]
\#\#\# Task description \\
An instruction (may include conversation history), a response to evaluate, and a score rubric are given. Follow these steps: \\
1. Evaluate the \textbf{\{judged score\}} of the response in detail strictly based on the score rubric. \\
2. Predict the score of the response from its evaluation. {\color{blue}The score is an integer between 0 and 4.} {\color{red}The score is an integer between 1 and 7.} \\
3. The output format should look as follows: "Evaluation: (detailed evaluation) Score: (predicted score)" \\
4. Do not generate any opening or closing statements. \\
\phantom{} \\
\#\#\# Instruction \\
\textbf{\{task\}} \\
\phantom{} \\
\#\#\# Response to evaluate \\
\textbf{\{response\}} \\
\phantom{} \\
\#\#\# Score rubric \\
\textbf{\{score rubrics\}} \\
\phantom{} \\
\#\#\# Hint \\
{\color{cyan}Evaluate the response as if its score was \textbf{\{conditioning\}}. Do not output the hint.} \\
\phantom{} \\
\#\#\# Output
  \end{tcolorbox}
  \caption{Our prompt and its posterior sampling variant. We set \textbf{\{task\}} to the evaluated task, \textbf{\{response\}} to its solution, \textbf{\{judged score\}} to the name of the metric, and \textbf{\{score rubrics\}} to its score rubrics. In prompt posterior sampling, we add the {\color{cyan}cyan text} and set \textbf{\{conditioning\}} to the human score. The {\color{blue}blue text} is only in the HelpSteer2 prompt, and the {\color{red}red text} is only in the Summarize from Feedback prompt.}
  \label{fig:prompt}
\end{figure}

\textbf{Prompt posterior sampling (\algpps).} Modern LLMs are instruction fine-tuned to follow human feedback \citep{ouyang22training}. Therefore, one natural way of implementing posterior sampling is by conditioning on the correct answer in the prompt, which we show in \cref{fig:prompt}. This directly addresses the computational inefficiency of \algrs. Specifically, because the correct answer is conditioned on, the model is likely to generate it and in this way increase the fraction of accepted rationale-answer pairs. We note that this does not \emph{guarantee} that the model outputs the correct answer. This also introduces a mismatch between training and testing. At training time, the LLM generates rationales by conditioning on both the question and answer, while at test time only the question is conditioned on. We denote the corresponding sampling distribution by $q_\textsc{pps}(\cdot \mid x, y^\star; \theta)$. As we show below, \algpps is essentially the second stage of \algstar.

\textbf{Self-taught reasoner sampling (\algstar).} Introduced by the seminal work of \citet{zelikman2022STaR}, \algstar has the following two stages:
\begin{enumerate}
  \item A single rejection sampling step (\algrs for $M = 1$).
  \item If the sampled answer is not correct, \algstar samples a rationale-answer pair using a rationalization prompt, with the correct answer as a hint. This is precisely our proposed \algpps.
\end{enumerate}
The pseudo-code of \algstar is given in \cref{alg:star-sampling}. Conceptually, \algstar chains \algrs and \algpps, effectively sampling rationales from a mixture of the two distributions. While this results in more accepted rationale-answer pairs on average per $\fem$ iteration than \algrs ($M = 1$) and \algpps, we show empirically that the increased quantity does not necessarily translate into quality. Instead, we demonstrate that the \emph{quality} of the sampling distribution $q$ is the decisive factor.

\section{Experiments}
\label{sec:experiments}

We evaluate \fem on \emph{LLM-as-a-judge calibration}~\citep{zheng23judging,sahoo2025llmjudge} and \emph{summarization from feedback}~\citep{stiennon2020summarize} tasks. We focus on these domains to highlight the impact of sampling distributions, which has been overlooked due to the community's major focus on math reasoning~\citep{cobbe2021gsm,math,shao24deepseekmath,lewkowycz2022solving,lightman2024lets,trinh2024alphageometry,luo2025wizardmath}. In complex math reasoning, conditioning on the final answer is not expected to provide much benefit because the tasks are complex. Our domains have unique properties that make them ideal for investigating rationale sampling schemes:
\begin{enumerate}
  \item \textbf{High information gain from $y^\star$.} The correct answer is a human preference score. The score is readily available in public datasets, and conditioning on it significantly improves the generated rationale.
  \item \textbf{Weak pre-trained models.} Pre-trained base models tend to perform poorly on our tasks because they lack knowledge about human scores in the domain, even if they are given the same score rubric as humans.
\end{enumerate}

\begin{figure*}[t!]
  \centering
  \includegraphics[width=5.6in,bb=0in 6in 7in 8in,clip]{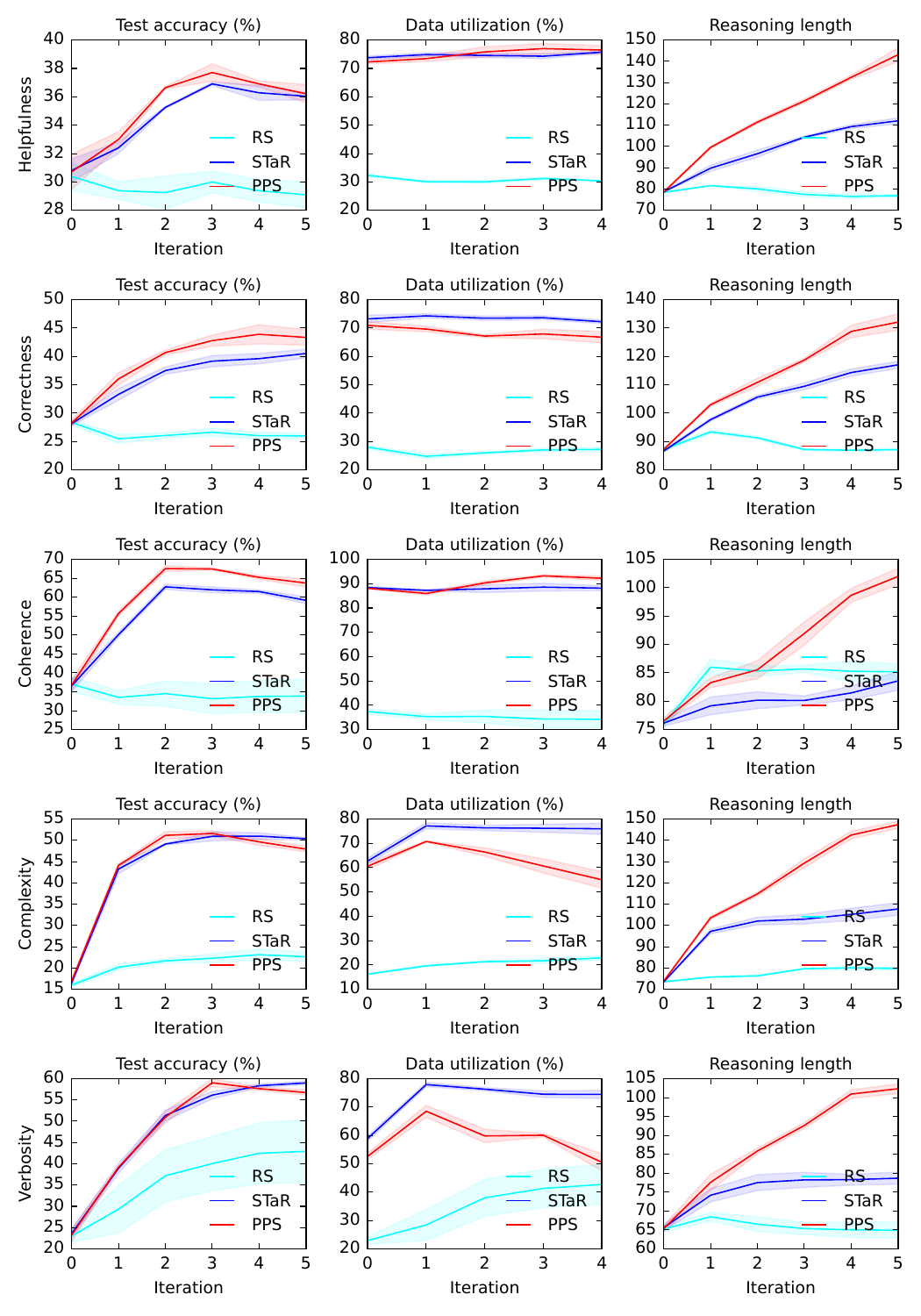}
  \caption{Test accuracy, data utilization, and reasoning length for Qwen 3B and HelpSteer2 correctness. The shaded areas are standard errors of the estimates from $3$ random runs.}
  \label{fig:run}
\end{figure*}

\begin{figure*}[t!]
  \centering
  \begin{tcolorbox}[title=Evaluation task,halign=left,fontupper=\small,fontlower=\small]
\textbf{Instruction:} Bob has two sons. John and Jay. Jay has one brother and father. Father has two sons. Jay's brother has a brother and a father. Who is Jay's brother? \textbf{Response:} Jay's brother is John.
  \end{tcolorbox}
  \begin{tcolorbox}[colframe=blue!50!black,colback=blue!10!white,title=Response before first iteration,halign=left,fontupper=\small,fontlower=\small]
\textbf{Evaluation:} The response is incorrect because it states that Jay's brother is John, while the instruction clearly indicates that Bob has two sons named John and Jay. Therefore, Jay's brother cannot be John. The response does not correctly address the question posed. \textbf{Score:} 0
  \end{tcolorbox}
  \begin{tcolorbox}[colframe=red!50!black,colback=red!10!white,title=Response after final iteration,halign=left,fontupper=\small,fontlower=\small]
\textbf{Evaluation:} The response accurately identifies Jay's brother as John based on the information provided in the prompt. The reasoning given aligns perfectly with the facts presented: Bob has two sons named John and Jay. Since Jay has one brother (John) and a father (Bob), and John also has a brother (Jay) and a father (Bob), it logically follows that Jay's brother must be John. There are no instances of incorrect information, misleading details, or hallucinations present in the response. The explanation clearly connects the relationships described in the prompt, making it a precise and correct representation of the scenario. Therefore, the response meets the criteria for a score of 4. \textbf{Score:} 4
  \end{tcolorbox}
  \caption{A qualitative example of a rationale improvement after learning with \algpps for Qwen 3B and HelpSteer2 correctness.}
  \label{tab:qualitative example}
\end{figure*}

\subsection{Setting}
\label{sec:experimental setting}

We experiment with two Llama \citep{Llama} (3B and 8B) and three Qwen \citep{Qwen2.5} (1.5B, 3B, and 7B) models on two tasks.\footnote{The specific models are Llama-3.2-3B-Instruct, Llama-3.1-8B-Instruct, Qwen2.5-1.5B-Instruct, Qwen2.5-3B-Instruct, and Qwen2.5-7B-Instruct from HuggingeFace.} The first task is calibration of an LLM judge against human scores. We experiment with \emph{HelpSteer2} dataset~\citep{wang2024helpsteer,wang2025helpsteerpreference}, a collection of instruction-response pairs rated on a $5$-point Likert scale. The quality of the responses is measured by $5$ metrics: helpfulness, correctness, coherence, complexity, and verbosity. The second task is predicting human quality scores for post summaries. We experiment with \emph{Summarize from Feedback} dataset~\citep{tldr,stiennon2020summarize}, a collection of Reddit post-summary pairs rated on a $7$-point Likert scale. We use its axis subset, which contains $4$ metrics: helpfulness, accuracy, coverage, and coherence. In summary, we evaluate $5$ models on predicting $9$ metrics in two tasks. Our main prompt is in \cref{fig:prompt}. The system prompt and score rubrics are presented in \cref{sec:prompts}.

We evaluate three $\fem$ sampling schemes: $\algrs$ with budget $M = 1$, $\algstar$, and $\algpps$; and compare them to pre-trained base models and GPT-5.2. The \textbf{E-step} in (Line 3 in \fem) is implemented using vLLM~\citep{vLLM} and the \textbf{M-step} (Line 5 in \fem) is implemented by SFT in TRL~\citep{vonwerra22trl}. We use AdamW~\citep{AdamW} with default parameters and one epoch in SFT. \fem is run for $K = 5$ iterations. The learning rate decays linearly from $\eta = 10^{-6}$ to $\eta = 10^{-7}$ across the iterations, and it is kept constant within each iteration. All experiments are conducted on NVIDIA A100 GPUs with resources scaled by model size: 1B models on 1 GPU, 3B on 2 GPUs, and 7-8B on 4 GPUs. All results are averaged over $3$ random runs with seeds $257$, $521$, and $1031$. We sample a random subset of $4 \,000$ data points ($2\,000$ train and $2\,000$ test) in each run. The sizes are chosen so that the initialization overhead of vLLM and SFT is negligible in the total runtime (less than $20\%$). Each run takes approximately $60$ minutes.

We report three metrics: test accuracy, data utilization, and reasoning length in words. The test accuracy is the accuracy in predicting human evaluation scores on a held-out test set, which we expect to improve as the number of \fem iterations increases. The data utilization is the percentage of accepted rationale-answer pairs, which are used for training. We report this number to show that more successful learning is not necessarily due to more training data. The reasoning length shows whether the improved performance is due to deeper or more compact reasoning.

\subsection{Main Results}
\label{sec:results}

\begin{table*}[t!]
  \centering
  \begin{minipage}[t]{0.49\textwidth}
    \caption{Highest test accuracies by all sampling schemes (\algpps is \colorbox{highlight}{shaded}) on all metrics in HelpSteer2 dataset.}
    \label{tab:hs2}
    \centering
    \resizebox{\linewidth}{!}{%
    {\small
    \begin{tabular}{ll|rrr>{\columncolor{highlight}}r} \hline
    Model & Metric & Base & \algrs & \algstar & \algpps \\ \hline
    Qwen & Helpfulness & 31.50 & 30.38 & \textbf{37.97} & 36.73 \\
    7B & Correctness & 26.56 & 24.05 & 40.50 & \textbf{43.73} \\
    & Coherence & 24.27 & 22.09 & 55.73 & \textbf{67.95} \\
    & Complexity & 28.84 & 44.71 & 53.03 & \textbf{53.99} \\
    & Verbosity & 15.56 & 17.97 & \textbf{61.90} & 61.85 \\ \hline
    Qwen & Helpfulness & 30.38 & 29.99 & 36.91 & \textbf{37.71} \\
    3B & Correctness & 28.38 & 26.64 & 40.51 & \textbf{43.91} \\
    & Coherence & 37.00 & 34.51 & 62.73 & \textbf{67.55} \\
    & Complexity & 15.99 & 23.12 & 50.98 & \textbf{51.60} \\
    & Verbosity & 23.02 & 42.92 & 58.99 & \textbf{59.04} \\ \hline
    Qwen & Helpfulness & 25.45 & 27.84 & 29.86 & \textbf{30.91} \\
    1B & Correctness & 23.04 & 23.60 & 32.78 & \textbf{33.01} \\
    & Coherence & 18.02 & 21.96 & 60.86 & \textbf{65.52} \\
    & Complexity & 15.22 & 18.23 & 45.48 & \textbf{47.99} \\
    & Verbosity & 23.62 & 38.02 & 55.42 & \textbf{55.63} \\ \hline
    Llama & Helpfulness & 34.86 & 36.87 & 36.83 & \textbf{37.63} \\
    8B & Correctness & 31.99 & 41.63 & 42.14 & \textbf{43.73} \\
    & Coherence & 45.27 & 68.99 & 67.22 & \textbf{71.42} \\
    & Complexity & 27.89 & 44.45 & 56.78 & \textbf{57.26} \\
    & Verbosity & 24.48 & 51.62 & 64.36 & \textbf{66.11} \\ \hline
    Llama & Helpfulness & 30.22 & 28.76 & 32.35 & \textbf{33.25} \\
    3B & Correctness & 21.55 & 19.28 & 39.71 & \textbf{39.73} \\
    & Coherence & 26.53 & 62.75 & \textbf{69.81} & 69.18 \\
    & Complexity & 23.97 & 28.50 & 54.40 & \textbf{54.66} \\
    & Verbosity & 38.26 & 50.24 & 61.50 & \textbf{61.71} \\ \hline
    \end{tabular}}
    }
  \end{minipage}
  \hfill %
  \begin{minipage}[t]{0.49\textwidth}
    \caption{Highest test accuracies by all sampling schemes (\algpps is \colorbox{highlight}{shaded}) on all metrics in Summarize from Feedback dataset.}
    \label{tab:sff}
    \centering
    \resizebox{\linewidth}{!}{%
    {\small
    \begin{tabular}{ll|rrr>{\columncolor{highlight}}r} \hline
    Model & Metric & Base & \algrs & \algstar & \algpps \\ \hline
    Qwen & Helpfulness & 17.85 & 18.76 & \textbf{22.61} & \textbf{22.61} \\
    7B & Accuracy & 8.54 & 10.35 & 67.32 & \textbf{71.37} \\
    & Coverage & 19.22 & 19.03 & 19.98 & \textbf{21.12} \\
    & Coherence & 5.14 & 5.26 & 71.32 & \textbf{73.04} \\ \hline
    Qwen & Helpfulness & 17.17 & 17.73 & 19.47 & \textbf{20.01} \\
    3B & Accuracy & 8.74 & 9.08 & 26.67 & \textbf{32.71} \\
    & Coverage & 12.91 & 14.16 & 15.40 & \textbf{16.54} \\
    & Coherence & 15.33 & 15.43 & 69.03 & \textbf{72.90} \\ \hline
    Qwen & Helpfulness & 13.86 & 14.56 & 20.18 & \textbf{20.64} \\
    1B & Accuracy & 22.83 & 36.93 & 58.24 & \textbf{66.14} \\
    & Coverage & 8.08 & 9.30 & 23.04 & \textbf{24.75} \\
    & Coherence & 17.27 & 20.11 & 74.51 & \textbf{76.26} \\ \hline
    Llama & Helpfulness & 13.96 & 16.34 & 20.56 & \textbf{20.93} \\
    8B & Accuracy & 9.89 & 15.42 & \textbf{73.26} & 72.50 \\
    & Coverage & 17.59 & 18.44 & 22.67 & \textbf{24.00} \\
    & Coherence & 11.41 & 61.70 & 71.17 & \textbf{74.22} \\ \hline
    Llama & Helpfulness & 14.14 & 18.61 & \textbf{23.73} & 22.19 \\
    3B & Accuracy & 11.38 & 14.21 & \textbf{69.49} & 69.20 \\
    & Coverage & 14.03 & 18.08 & \textbf{27.15} & 26.47 \\
    & Coherence & 5.33 & 7.44 & 71.33 & \textbf{76.66} \\ \hline
    \end{tabular}}
    }
  \end{minipage}
\end{table*}

Our results for Qwen 3B and HelpSteer2 correctness are reported in \cref{fig:run}. The plots for all other models and metrics are reported in \cref{sec:detailed fem runs}. We observe two key trends. First, the test accuracies of \algstar and \algpps increase with more iterations and plateau, confirming the effectiveness of multiple \fem iterations. Second, \algpps outperforms all baselines across all \fem iterations, which we attribute to generating superior rationales.

Crucially, the data utilization shows that this performance gain is not driven by data quantity; \algpps achieves a higher accuracy with less training data than \algstar. Instead, it is driven by the quality of reasoning. The reasoning length of \algpps increases to $130$ words versus less than $120$ for \algstar. Qualitatively, the longer reasoning is more comprehensive, and we illustrate it in \cref{tab:qualitative example}. The final rationale resolves confusion in the original rationale, which says that Jay and John cannot be brothers because they have the same father. This mirrors observations in RL-based training~\citep{deepseek-r1} where models learn to reason more deeply.

We want to stress that this observed trend is model- and domain-dependent, as can be seen in \cref{sec:detailed fem runs}. In some cases, the reasoning length \emph{decreases} while accuracy improves. This suggests that \fem does not blindly encourage reasoning length, but rather learns to reason in a way that is most suitable for the specific task and model: deeper reasoning for complex judgments, or improving conciseness and compactness of the language.

We summarize the highest accuracies attained by all sampling schemes in \cref{tab:hs2,tab:sff}. In HelpSteer2 experiments (\cref{tab:hs2}), \algpps has the highest accuracy in $22$ out of $25$ and \algstar in $3$. Under the null hypothesis that \algpps and \algstar perform identically, the p-value of winning at least $22$ times is less than $0.01$. In Summarize from Feedback experiments (\cref{tab:sff}), \algpps has the highest accuracy in $15$ out of $20$, \algstar in $4$, and they tie in one. Under the null hypothesis that \algpps and \algstar perform identically, the p-value of winning at least $15$ times is about $0.02$. We conclude that \algpps outperforms the closest baseline with a high statistical significance.

To further show that our tasks are non-trivial for state-of-the-art LLMs, we evaluate GPT-5.2 as a strong baseline. In HelpSteer2 dataset, GPT-5.2 achieves accuracies $32.91\%$ (helpfulness), $30.84\%$ (correctness), $65.08\%$ (coherence), $58.88\%$ (complexity), and $43.19\%$ (verbosity). Our best fine-tuned models surpass these numbers in all metrics but complexity. In Summarize from Feedback dataset, GPT-5.2 achieves $12.40\%$ (helpfulness), $69.91\%$ (accuracy), $16.40\%$ (coverage), and $50.43\%$ (coherence). Notably, our best fine-tuned models surpass all of these. This confirms that our tasks genuinely benefit from adaptation to the domain.

\subsection{Ablation Studies}

We investigate the robustness of our findings through three ablation studies with Qwen 3B (\cref{sec:ablations studies}). First, we vary the rejection sampling budget $M \in \{1, 3, 5\}$ in \algrs. While increasing $M$ sometimes improves performance through increased data utilization, it also increases computation time linearly in $M$, and still significantly underperforms \algpps and \algstar. Second, we test the sensitivity of our results to the prompt in \cref{fig:prompt} by conditioning less explicitly. We observe that less precise conditioning decreases absolute performance of all sampling schemes. However, the relative order remains the same, and \algpps continues to outperform all baselines. This shows that our results generalize beyond a single prompt and highlights the importance of precise instructions when conditioning. Finally, we also experiment with smaller and larger training set sizes.

\section{Related Work}
\label{sec:related work}

Our filtered EM (\fem) framework unifies several recent approaches to self-improvement and reasoning under the lens of inference. In this section, we explicate the connections to specific EM-style algorithms, data filtering strategies, and the broader RL as inference literature.

\textbf{LLM self-improvement.} \fem encompasses several recent self-improvement algorithms for LLM reasoning. Notably, \textit{ReSTEM}~\citep{singh2024ReSTEM} correspond to instantiating the rationale proposal distribution $q$ in Algorithm~\ref{alg:EM-filter} with standard \emph{rejection sampling}~\citep{neal1998EM}. This relationship arises from a specific modeling choice: following the variational perspective of \citet{dayan-hinton}, \citet{singh2024ReSTEM} introduce a binary optimality variable $\mathcal{O}$ and maximize $\sP(\mathcal{O}=1 \mid x)$, treating the entire trajectory—rationale $z$ and answer $y$—as a latent variable. Consequently, their \textbf{E-step} filters samples from a fixed distribution: the LLM conditioned \emph{only} on the question $x$. Such reliance on a fixed sampling distribution persists in recent literature as well. For instance, \textit{LaTRO}~\citep{chen2024latro} introduces a variational framework optimizing a lower bound on correct answer likelihood; \citet{xu2025EM} proposes \textit{EM Policy Gradient} to stabilize off-policy updates; and \textit{RAFT++}~\citep{xiong2025RAFT++} proposes a filtered version of PPO~\citep{PPO}. In all these cases, the algorithms operate under a fixed sampling policy $\pi(z, y \mid x; \theta)$ that conditions only on the question.

Other works have explored alternative proposal distributions, but typically through complex probabilistic machinery or separate inference networks. \textit{TRICE}~\citep{phan2023TRICE} addresses the high variance of gradient estimators in the \textbf{M-step} using control variates. \textit{BRiTE}~\citep{zhong2025BRiTE} targets the \textbf{E-step} by jointly training a separate LLM to approximate the posterior via reinforcement learning. Similarly, \citet{hu2024amortizing} propose amortizing intractable posterior inference using GFlowNets~\citep{bengio2023gflownet}; although they do not apply this to iterative LLM training, it shares the limitation of requiring a separate trained network to sample from the true posterior.

Our \fem explores a more lightweight alternative: leveraging the LLM's inherent reasoning and instruction-following capabilities~\citep{bai2022instruction-following,ouyang22training,CoT,CoT2} to approximate $q$ via prompting, thereby avoiding the complexity of training auxiliary models. We demonstrate that in notable domains, such as calibrating LLM-as-a-judge, this EM-based approach can be vastly improved by simply changing the sampling distribution to that of rationales conditioned on the true answer. This highlights that, as prior work has overlooked, examining the sampling scheme offers a simple yet effective direction to further improve the training efficiency of LLM reasoning.

\textbf{Filtering and implicit objectives.} Beyond iterative algorithms, our work connects to the literature, viewing data filtration as implicit objective modification. \citet{shrivastava2025gfpo} propose oversampling reasoning traces and applying top-$K$ filtering, arguing that such filtration acts as a form of reward shaping. Similarly, \citet{karan2025base} shows that inference-time sampling from a sharpened distribution can outperform RL-based post-training. From a distributional perspective, methods like \textit{Quark}~\citep{khalifa2021dm} and others~\citep{go2023f-div,kruszewski2025filtering,kim2025guaranteed} derive objectives by defining a target distribution that discards incorrect answers while preserving the relative probabilities of correct ones, often approximated via information projection~\citep{amari1985,nielsen2018information}.

Our work is complementary to this line of research: we adopt a latent-variable perspective, treating such filtering as an implicit target distribution over rationales and answers, and study how different choices of the \emph{sampling} distribution in the \textbf{E-step} affect learning via our \fem. Rather than viewing filtering as a heuristic for dataset cleaning, \fem frames it as a formal approximation of the posterior in a latent variable model.

\textbf{RL as probabilistic inference.} Finally, the derivation of \fem is grounded in the probabilistic formulation of RL, also known as Control as Inference; see our discussions with Lemma~\ref{lem:mukherjee}. The paradigm of addressing RL problems via Expectation-Maximization is well established in control theory~\citep{dayan-hinton,levine2018survey}, having been investigated through various lenses, including mixture models~\citep{vlassis2009EM} and MCMC approximations~\citep{hoffman2007mcmc}. While this theoretical equivalence is well-documented in general RL, our work systematically operationalizes it for LLM reasoning. We underscore that, in the context of LLMs, the choice of sampling distribution is not merely a heuristic choice but a central component of the variational objective.

\section{Conclusions}
\label{sec:conclusions}

We revisit reasoning as a latent-variable model, bridging the gap between learning to reason, filtered EM, and reward-based policy gradients. Our work highlights the critical role of rationale sampling distributions, a factor that may have a major impact on empirical performance. In the process, we propose prompt posterior sampling, a simple yet effective sampling scheme corresponding to the rationalization stage of \algstar. In our experiments, prompt posterior sampling consistently outperforms other sampling schemes, including rejection sampling and \algstar, on two human-rating alignment tasks in HelpSteer2 and Summarize from Feedback datasets.

\textbf{Limitations.} First, we establish that the sampling distribution is an important domain-specific hyperparameter. Our results on alignment tasks provide a definitive proof of concept: the \say{right} choice of the rationale sampling distribution in \fem leads to substantial empirical gains. In future work, we plan to experiment with more complex tasks, such as math reasoning, and to explore richer guiding signals. Second, our current algorithm design is motivated by SFT, in which each iteration is essentially a single epoch during which the training data are adapted to the learned model using sampled rationales. In future work, we plan to investigate if our improvements in sampling generalize to online reinforcement learning, such as GRPO \citep{shao24deepseekmath}. When viewed in this way, our current design is reminiscent of batch RL \citep{lange12batch,jaques20humancentric,levine2018survey}, also with connections to prior literature on expert iterations~\citep{anthony2017ExpertIteration,anthony2021thesis} Last but not least, while all of our approximations in \cref{sec:em to fine-tuning} are properly justified, we have not formally analyzed \fem, which we also leave for future work. This is because we focused on developing the algorithmic framework and empirically validating it.

\section*{Impact Statement}

This paper presents work whose goal is to advance the field of machine learning. There are many potential societal consequences of our work, none of which we feel must be specifically highlighted here.

\clearpage

\bibliographystyle{icml2026}
\bibliography{bib/Brano,bib/Junghyun}

@book{ doucet01sequential,
  author = "Arnaud Doucet and Nando de Freitas and Neil Gordon",
  title = "Sequential {Monte Carlo} Methods in Practice",
  publisher = "Springer",
  address = "New York, NY",
  year = "2001"
}

@article{cappe09online,
    author = {Capp\'{e}, Olivier and Moulines, Eric},
    title = {{On-Line Expectation–Maximization Algorithm for latent Data Models}},
    journal = {Journal of the Royal Statistical Society Series B: Statistical Methodology},
    volume = {71},
    number = {3},
    pages = {593-613},
    year = {2009},
    month = {02},
    issn = {1369-7412},
    doi = {10.1111/j.1467-9868.2009.00698.x},
}

@inbook{ lange12batch,
  author = "Sascha Lange and Thomas Gabel and Martin Riedmiller",
  title = "Batch Reinforcement Learning",
  booktitle = "Reinforcement Learning: State-of-the-Art",
  pages = "45-73",
  publisher = "Springer Berlin Heidelberg",
  address = "Berlin, Heidelberg",
  year = "2012"
}

@inproceedings{ ma19imitationregularized,
  author = "Yifei Ma and Yu-Xiang Wang and Balakrishnan Narayanaswamy",
  title = "Imitation-Regularized Offline Learning",
  booktitle = "Proceedings of the 22nd International Conference on Artificial Intelligence and Statistics",
  year = "2019"
}

@inproceedings{ jaques20humancentric,
  author = "Natasha Jaques and Judy Hanwen Shen and Asma Ghandeharioun and Craig Ferguson and Agata Lapedriza and Noah Jones and Shixiang Gu and Rosalind Picard",
  title = "Human-Centric Dialog Training via Offline Reinforcement Learning",
  booktitle = "Proceedings of the 2020 Conference on Empirical Methods in Natural Language Processing",
  year = "2020"
}

@inproceedings{ouyang22training,
 author = {Ouyang, Long and Wu, Jeffrey and Jiang, Xu and Almeida, Diogo and Wainwright, Carroll and Mishkin, Pamela and Zhang, Chong and Agarwal, Sandhini and Slama, Katarina and Ray, Alex and Schulman, John and Hilton, Jacob and Kelton, Fraser and Miller, Luke and Simens, Maddie and Askell, Amanda and Welinder, Peter and Christiano, Paul F and Leike, Jan and Lowe, Ryan},
 booktitle = {Advances in Neural Information Processing Systems},
 pages = {27730--27744},
 publisher = {Curran Associates, Inc.},
 title = {Training language models to follow instructions with human feedback},
 url = {https://openreview.net/forum?id=TG8KACxEON},
 volume = {35},
 year = {2022}
}

@inproceedings{ zheng23judging,
  author = "Lianmin Zheng and Wei-Lin Chiang and Ying Sheng and Siyuan Zhuang and Zhanghao Wu and Yonghao Zhuang and Zi Lin and Zhuohan Li and Dacheng Li and Eric Xing and Hao Zhang and Joseph Gonzalez and Ion Stoica",
  title = "Judging {LLM}-as-a-Judge with {MT-Bench} and {Chatbot Arena}",
  booktitle = "Advances in Neural Information Processing Systems 36",
  year = "2023"
}

@article{shao24deepseekmath,
    title={{DeepSeekMath: Pushing the Limits of Mathematical Reasoning in Open Language Models}}, 
    author={Zhihong Shao and Peiyi Wang and Qihao Zhu and Runxin Xu and Junxiao Song and Xiao Bi and Haowei Zhang and Mingchuan Zhang and Y. K. Li and Y. Wu and Daya Guo},
    year={2024},
    journal={arXiv preprint arXiv:2402.03300},
    url={https://arxiv.org/abs/2402.03300}, 
}

@inproceedings{ ankner24critiqueoutloud,
  author = "Zachary Ankner and Mansheej Paul and Brandon Cui and Jonathan Daniel Chang and Prithviraj Ammanabrolu",
  title = "Critique-out-Loud Reward Models",
  booktitle = "Pluralistic Alignment Workshop at NeurIPS 2024",
  year = "2024"
}

@inproceedings{chiang25tract,
    title = {{TRACT: Regression-Aware Fine-tuning Meets Chain-of-Thought Reasoning for LLM-as-a-Judge}},
    author = "Chiang, Cheng-Han  and
      Lee, Hung-yi  and
      Lukasik, Michal",
    booktitle = "Proceedings of the 63rd Annual Meeting of the Association for Computational Linguistics (Volume 1: Long Papers)",
    month = jul,
    year = "2025",
    address = "Vienna, Austria",
    publisher = "Association for Computational Linguistics",
    doi = "10.18653/v1/2025.acl-long.147",
    pages = "2934--2952",
    ISBN = "979-8-89176-251-0",
}

@misc{ vonwerra22trl,
  author = "Leandro von Werra and Younes Belkada and Lewis Tunstall and Edward Beeching and Tristan Thrush and Nathan Lambert and Shengyi Huang and Kashif Rasul and Quentin Gallouedec",
  title = "{TRL: Transformer Reinforcement Learning}",
  howpublished = "\url{https://github.com/huggingface/trl}",
  year = "2020",
}

@inproceedings{anthony2017ExpertIteration,
 author = {Anthony, Thomas and Tian, Zheng and Barber, David},
 booktitle = {Advances in Neural Information Processing Systems},
 pages = {5366--5376},
 publisher = {Curran Associates, Inc.},
 title = {{Thinking Fast and Slow with Deep Learning and Tree Search}},
 url = {https://proceedings.neurips.cc/paper_files/paper/2017/file/d8e1344e27a5b08cdfd5d027d9b8d6de-Paper.pdf},
 volume = {30},
 year = {2017}
}

@phdthesis{anthony2021thesis,
  author       = {Anthony, Thomas William},
  title        = {{Expert Iteration}},
  school       = {UCL (University College London)},
  year         = {2021},
  type         = {Ph.D. thesis},
  url          = {https://discovery.ucl.ac.uk/id/eprint/10123580/},
  language     = {English}
}

@inproceedings{zelikman2022STaR,
 author = {Zelikman, Eric and Wu, Yuhuai and Mu, Jesse and Goodman, Noah},
 booktitle = {Advances in Neural Information Processing Systems},
 pages = {15476--15488},
 publisher = {Curran Associates, Inc.},
 title = {{STaR: Bootstrapping Reasoning With Reasoning}},
 url = {https://openreview.net/forum?id=_3ELRdg2sgI},
 volume = {35},
 year = {2022}
}

@article{yuan2023RFT,
  title={{Scaling Relationship on Learning Mathematical Reasoning with Large Language Models}},
  author={Yuan, Zheng and Yuan, Hongyi and Li, Chengpeng and Dong, Guanting and Lu, Keming and Tan, Chuanqi and Zhou, Chang and Zhou, Jingren},
  journal={arXiv preprint arXiv:2308.01825},
  year={2023},
  url={https://arxiv.org/abs/2308.01825}
}

@article{liang2022local,
  title={{Local Policy Improvement for Recommender Systems}},
  author={Dawen Liang and Nikos Vlassis},
  journal={arXiv preprint arXiv:2212.11431},
  year={2022},
  url={https://arxiv.org/abs/2212.11431}
}

@article{dong2023RAFT,
title={{RAFT: Reward rAnked FineTuning for Generative Foundation Model Alignment}},
author={Hanze Dong and Wei Xiong and Deepanshu Goyal and Yihan Zhang and Winnie Chow and Rui Pan and Shizhe Diao and Jipeng Zhang and KaShun Shum and Tong Zhang},
journal={Transactions on Machine Learning Research},
issn={2835-8856},
year={2023},
url={https://openreview.net/forum?id=m7p5O7zblY},
}

@article{xiong2025RAFT++,
    title={{A Minimalist Approach to LLM Reasoning: from Rejection Sampling to Reinforce}}, 
    author={Wei Xiong and Jiarui Yao and Yuhui Xu and Bo Pang and Lei Wang and Doyen Sahoo and Junnan Li and Nan Jiang and Tong Zhang and Caiming Xiong and Hanze Dong},
    year={2025},
    journal={arXiv preprint arXiv:2504.11343},
    url={https://arxiv.org/abs/2504.11343}, 
}

@inproceedings{phan2023TRICE,
 author = {Phan, Du and Hoffman, Matthew Douglas and Dohan, David and Douglas, Sholto and Le, Tuan Anh and Parisi, Aaron and Sountsov, Pavel and Sutton, Charles and Vikram, Sharad and A. Saurous, Rif},
 booktitle = {Advances in Neural Information Processing Systems},
 pages = {72819--72841},
 publisher = {Curran Associates, Inc.},
 title = {{Training Chain-of-Thought via Latent-Variable Inference}},
 url = {https://openreview.net/forum?id=a147pIS2Co},
 volume = {36},
 year = {2023}
}

@article{singh2024ReSTEM,
title={{Beyond Human Data: Scaling Self-Training for Problem-Solving with Language Models}},
author={Avi Singh and John D Co-Reyes and Rishabh Agarwal and Ankesh Anand and Piyush Patil and Xavier Garcia and Peter J Liu and James Harrison and Jaehoon Lee and Kelvin Xu and Aaron T Parisi and Abhishek Kumar and Alexander A Alemi and Alex Rizkowsky and Azade Nova and Ben Adlam and Bernd Bohnet and Gamaleldin Fathy Elsayed and Hanie Sedghi and Igor Mordatch and Isabelle Simpson and Izzeddin Gur and Jasper Snoek and Jeffrey Pennington and Jiri Hron and Kathleen Kenealy and Kevin Swersky and Kshiteej Mahajan and Laura A Culp and Lechao Xiao and Maxwell Bileschi and Noah Constant and Roman Novak and Rosanne Liu and Tris Warkentin and Yamini Bansal and Ethan Dyer and Behnam Neyshabur and Jascha Sohl-Dickstein and Noah Fiedel},
journal={Transactions on Machine Learning Research},
issn={2835-8856},
year={2024},
url={https://openreview.net/forum?id=lNAyUngGFK},
note={Expert Certification}
}

@inproceedings{hu2024amortizing,
title={{Amortizing intractable inference in large language models}},
author={Edward J Hu and Moksh Jain and Eric Elmoznino and Younesse Kaddar and Guillaume Lajoie and Yoshua Bengio and Nikolay Malkin},
booktitle={The Twelfth International Conference on Learning Representations},
year={2024},
url={https://openreview.net/forum?id=Ouj6p4ca60}
}

@article{bengio2023gflownet,
  author  = {Yoshua Bengio and Salem Lahlou and Tristan Deleu and Edward J. Hu and Mo Tiwari and Emmanuel Bengio},
  title   = {{GFlowNet Foundations}},
  journal = {Journal of Machine Learning Research},
  year    = {2023},
  volume  = {24},
  number  = {210},
  pages   = {1--55},
  url     = {http://jmlr.org/papers/v24/22-0364.html}
}

@InProceedings{zhong2025BRiTE,
	title = 	 {{BRiTE: Bootstrapping Reinforced Thinking Process to Enhance Language Model Reasoning}},
	author =       {Han Zhong and Yutong Yin and Shenao Zhang and Xiaojun Xu and Yuanxin Liu and Yifei Zuo and Zhihan Liu and Boyi Liu and Sirui Zheng and Hongyi Guo and Liwei Wang and Mingyi Hong and Zhaoran Wang},
	booktitle = 	 {Proceedings of the 42nd International Conference on Machine Learning},
	pages = 	 {},
	year = 	 {2025},
	editor = 	 {},
	volume = 	 {},
	series = 	 {Proceedings of Machine Learning Research},
	month = 	 {13--19 Jul},
	publisher =    {PMLR},
	url = 	 {https://openreview.net/forum?id=NME3HKUHLX},
}

@article{tang2025JEPO,
    title={{Beyond Verifiable Rewards: Scaling Reinforcement Learning for Language Models to Unverifiable Data}}, 
    author={Yunhao Tang and Sid Wang and Lovish Madaan and Rémi Munos},
    year={2025},
    journal={arXiv preprint arXiv:2503.19618},
    url={https://arxiv.org/abs/2503.19618}, 
}

@article{xu2025EM,
    title={{Training Large Language Models to Reason via EM Policy Gradient}},
    author={Tianbing Xu},
    journal={arXiv preprint arXiv:2504.18587},
    url={https://arxiv.org/abs/2504.18587},
    year={2025}
}

@misc{PPO,
    title={{Proximal Policy Optimization Algorithms}}, 
    author={John Schulman and Filip Wolski and Prafulla Dhariwal and Alec Radford and Oleg Klimov},
    year={2017},
    journal={arXiv preprint arXiv:1707.06347},
    url={https://arxiv.org/abs/1707.06347}, 
}

@article{deepseek-r1,
      title={{DeepSeek-R1: Incentivizing Reasoning Capability in LLMs via Reinforcement Learning}}, 
      author={DeepSeek-AI},
      year={2025},
      journal={arXiv preprint arXiv:2501.12948},
      url={https://arxiv.org/abs/2501.12948}, 
}

@article{shrivastava2025gfpo,
      title={{Sample More to Think Less: Group Filtered Policy Optimization for Concise Reasoning}}, 
      author={Vaishnavi Shrivastava and Ahmed Awadallah and Vidhisha Balachandran and Shivam Garg and Harkirat Behl and Dimitris Papailiopoulos},
      year={2025},
      journal={arXiv preprint arXiv:2508.09726},
      url={https://arxiv.org/abs/2508.09726}, 
}

@article{chen2024latro,
    title={{Language Models are Hidden Reasoners: Unlocking Latent Reasoning Capabilities via Self-Rewarding}}, 
    author={Haolin Chen and Yihao Feng and Zuxin Liu and Weiran Yao and Akshara Prabhakar and Shelby Heinecke and Ricky Ho and Phil Mui and Silvio Savarese and Caiming Xiong and Huan Wang},
    year={2024},
    journal={arXiv preprint arXiv:2411.04282},
    url={https://arxiv.org/abs/2411.04282}, 
}

@inproceedings{CoT,
 author = {Wei, Jason and Wang, Xuezhi and Schuurmans, Dale and Bosma, Maarten and ichter, brian and Xia, Fei and Chi, Ed and Le, Quoc V and Zhou, Denny},
 booktitle = {Advances in Neural Information Processing Systems},
 pages = {24824--24837},
 publisher = {Curran Associates, Inc.},
 title = {Chain-of-Thought Prompting Elicits Reasoning in Large Language Models},
 url = {https://openreview.net/forum?id=_VjQlMeSB_J},
 volume = {35},
 year = {2022}
}

@inproceedings{CoT2,
 author = {Kojima, Takeshi and Gu, Shixiang (Shane) and Reid, Machel and Matsuo, Yutaka and Iwasawa, Yusuke},
 booktitle = {Advances in Neural Information Processing Systems},
 pages = {22199--22213},
 publisher = {Curran Associates, Inc.},
 title = {{Large Language Models are Zero-Shot Reasoners}},
 url = {https://openreview.net/forum?id=e2TBb5y0yFf},
 volume = {35},
 year = {2022}
}

@article{bai2022instruction-following,
    title={{Training a Helpful and Harmless Assistant with Reinforcement Learning from Human Feedback}}, 
    author={Yuntao Bai and Andy Jones and Kamal Ndousse and Amanda Askell and Anna Chen and Nova DasSarma and Dawn Drain and Stanislav Fort and Deep Ganguli and Tom Henighan and Nicholas Joseph and Saurav Kadavath and Jackson Kernion and Tom Conerly and Sheer El-Showk and Nelson Elhage and Zac Hatfield-Dodds and Danny Hernandez and Tristan Hume and Scott Johnston and Shauna Kravec and Liane Lovitt and Neel Nanda and Catherine Olsson and Dario Amodei and Tom Brown and Jack Clark and Sam McCandlish and Chris Olah and Ben Mann and Jared Kaplan},
    year={2022},
    journal={arXiv preprint arXiv:2204.05862},
    url={https://arxiv.org/abs/2204.05862}, 
}

@article{karan2025base,
      title={{Reasoning with Sampling: Your Base Model is Smarter Than You Think}}, 
      author={Aayush Karan and Yilun Du},
      year={2025},
      journal={arXiv preprint arXiv:2510.14901},
      url={https://arxiv.org/abs/2510.14901}, 
}

@inproceedings{mukherjee2025multiturn,
 title={{Offline RL by Reward-Weighted Fine-Tuning for Conversation Optimization}}, 
 author={Subhojyoti Mukherjee and Viet Dac Lai and Raghavendra Addanki and Ryan Rossi and Seunghyun Yoon and Trung Bui and Anup Rao and Jayakumar Subramanian and Branislav Kveton},
 booktitle = {Advances in Neural Information Processing Systems},
 publisher = {Curran Associates, Inc.},
 url = {https://openreview.net/forum?id=WAFD6VYIEa},
 volume = {38},
 year = {2025}
}

@article{dempster1977EM,
    author = {Dempster, A. P. and Laird, N. M. and Rubin, D. B.},
    title = {{Maximum Likelihood from Incomplete Data Via the EM Algorithm}},
    journal = {Journal of the Royal Statistical Society: Series B (Methodological)},
    volume = {39},
    number = {1},
    pages = {1-22},
    year = {1977},
    month = {12},
    issn = {0035-9246},
    doi = {10.1111/j.2517-6161.1977.tb01600.x},
}

@article{MCEM,
author = {Greg C. G. Wei and Martin A. Tanner},
title = {{A Monte Carlo Implementation of the EM Algorithm and the Poor Man's Data Augmentation Algorithms}},
journal = {Journal of the American Statistical Association},
volume = {85},
number = {411},
pages = {699--704},
year = {1990},
publisher = {ASA Website},
doi = {10.1080/01621459.1990.10474930},
}

@incollection{neath2013convergence,
  author    = {Neath, Ronald C.},
  title     = {{On Convergence Properties of the Monte Carlo EM Algorithm}},
  booktitle = {Advances in Modern Statistical Theory and Applications: A Festschrift in Honor of Morris L. Eaton},
  editor    = {Jones, Galin and Shen, Xiaotong},
  publisher = {Institute of Mathematical Statistics},
  year      = {2013},
  series    = {Institute of Mathematical Statistics Collections},
  volume    = {10},
  pages     = {43--62},
  doi       = {10.1214/12-IMSCOLL1003},
}

@article{dayan-hinton,
    author = {Dayan, Peter and Hinton, Geoffrey E.},
    title = {{Using Expectation-Maximization for Reinforcement Learning}},
    journal = {Neural Computation},
    volume = {9},
    number = {2},
    pages = {271-278},
    year = {1997},
    month = {02},
    issn = {0899-7667},
    doi = {10.1162/neco.1997.9.2.271},
}

@inproceedings{hoffman2007mcmc,
 author = {Hoffman, Matthew and Doucet, Arnaud and Freitas, Nando and Jasra, Ajay},
 booktitle = {Advances in Neural Information Processing Systems},
 pages = {},
 publisher = {Curran Associates, Inc.},
 title = {{Bayesian Policy Learning with Trans-Dimensional MCMC}},
 url = {https://proceedings.neurips.cc/paper_files/paper/2007/file/3a15c7d0bbe60300a39f76f8a5ba6896-Paper.pdf},
 volume = {20},
 year = {2007}
}

@ARTICLE{moon1996EM,
  author={Moon, T.K.},
  journal={IEEE Signal Processing Magazine}, 
  title={The expectation-maximization algorithm}, 
  year={1996},
  volume={13},
  number={6},
  pages={47-60},
  doi={10.1109/79.543975}
}

@Inbook{neal1998EM,
author="Neal, Radford M.
and Hinton, Geoffrey E.",
title={{A View of the EM Algorithm that Justifies Incremental, Sparse, and other Variants}},
bookTitle="Learning in Graphical Models",
year="1998",
publisher="Springer Netherlands",
address="Dordrecht",
pages="355--368",
isbn="978-94-011-5014-9",
doi="10.1007/978-94-011-5014-9_12",
}

@article{balakrishnan2017EM,
author = {Sivaraman Balakrishnan and Martin J. Wainwright and Bin Yu},
title = {{Statistical guarantees for the EM algorithm: From population to sample-based analysis}},
volume = {45},
journal = {The Annals of Statistics},
number = {1},
publisher = {Institute of Mathematical Statistics},
pages = {77 -- 120},
year = {2017},
doi = {10.1214/16-AOS1435},
}

@inproceedings{vlassis2009EM,
author = {Vlassis, Nikos and Toussaint, Marc},
title = {{Model-Free Reinforcement Learning as Mixture Learning}},
year = {2009},
isbn = {9781605585161},
publisher = {Association for Computing Machinery},
address = {New York, NY, USA},
url = {https://orbilu.uni.lu/bitstream/10993/3376/1/09-vlassis-toussaint-ICML.pdf},
doi = {10.1145/1553374.1553512},
booktitle = {Proceedings of the 26th Annual International Conference on Machine Learning},
pages = {1081–1088},
numpages = {8},
location = {Montreal, Quebec, Canada},
series = {ICML '09}
}

@article{levine2018survey,
      title={{Reinforcement Learning and Control as Probabilistic Inference: Tutorial and Review}}, 
      author={Sergey Levine},
      year={2018},
      journal={arXiv preprint arXiv:1805.00909},
      url={https://arxiv.org/abs/1805.00909}, 
}

@article{cobbe2021gsm,
      title={{Training Verifiers to Solve Math Word Problems}}, 
      author={Karl Cobbe and Vineet Kosaraju and Mohammad Bavarian and Mark Chen and Heewoo Jun and Lukasz Kaiser and Matthias Plappert and Jerry Tworek and Jacob Hilton and Reiichiro Nakano and Christopher Hesse and John Schulman},
      year={2021},
      journal={arXiv preprint arXiv:2110.14168},
      url={https://arxiv.org/abs/2110.14168}, 
}

@inproceedings{
lewkowycz2022solving,
title={{Solving Quantitative Reasoning Problems with Language Models}},
author={Aitor Lewkowycz and Anders Johan Andreassen and David Dohan and Ethan Dyer and Henryk Michalewski and Vinay Venkatesh Ramasesh and Ambrose Slone and Cem Anil and Imanol Schlag and Theo Gutman-Solo and Yuhuai Wu and Behnam Neyshabur and Guy Gur-Ari and Vedant Misra},
booktitle={Advances in Neural Information Processing Systems},
year={2022},
url={https://openreview.net/forum?id=IFXTZERXdM7}
}

@inproceedings{
lightman2024lets,
title={{Let's Verify Step by Step}},
author={Hunter Lightman and Vineet Kosaraju and Yuri Burda and Harrison Edwards and Bowen Baker and Teddy Lee and Jan Leike and John Schulman and Ilya Sutskever and Karl Cobbe},
booktitle={The Twelfth International Conference on Learning Representations},
year={2024},
url={https://openreview.net/forum?id=v8L0pN6EOi}
}

@article{trinh2024alphageometry,
author={Trinh, Trieu H. and Wu, Yuhuai and Le, Quoc V. and He, He and Luong, Thang},
title={Solving olympiad geometry without human demonstrations},
journal={Nature},
year={2024},
month={Jan},
day={01},
volume={625},
number={7995},
pages={476-482},
issn={1476-4687},
doi={10.1038/s41586-023-06747-5},
}

@inproceedings{
luo2025wizardmath,
title={{WizardMath: Empowering Mathematical Reasoning for Large Language Models via Reinforced Evol-Instruct}},
author={Haipeng Luo and Qingfeng Sun and Can Xu and Pu Zhao and Jian-Guang Lou and Chongyang Tao and Xiubo Geng and Qingwei Lin and Shifeng Chen and Yansong Tang and Dongmei Zhang},
booktitle={The Thirteenth International Conference on Learning Representations},
year={2025},
url={https://openreview.net/forum?id=mMPMHWOdOy}
}

@article{Llama,
  title={{The Llama 3 Herd of Models}},
  author={{Llama Team}},
  journal={arXiv preprint arXiv:2407.21783},
  year={2024},
  url={https://arxiv.org/abs/2407.21783}
}

@article{Qwen2.5,
  title={{Qwen2.5 Technical Report}},
  author={{Qwen Team}},
  journal={arXiv preprint arXiv:2412.15115},
  year={2024},
  url={https://arxiv.org/abs/2412.15115}
}

@inproceedings{wang2024helpsteer,
title={{HelpSteer 2: Open-source dataset for training top-performing reward models}},
author={Zhilin Wang and Yi Dong and Olivier Delalleau and Jiaqi Zeng and Gerald Shen and Daniel Egert and Jimmy J. Zhang and Makesh Narsimhan Sreedhar and Oleksii Kuchaiev},
booktitle={The Thirty-eight Conference on Neural Information Processing Systems Datasets and Benchmarks Track},
year={2024},
url={https://openreview.net/forum?id=PvVKUFhaNy}
}

@inproceedings{wang2025helpsteerpreference,
title={{HelpSteer2-Preference: Complementing Ratings with Preferences}},
author={Zhilin Wang and Alexander Bukharin and Olivier Delalleau and Daniel Egert and Gerald Shen and Jiaqi Zeng and Oleksii Kuchaiev and Yi Dong},
booktitle={The Thirteenth International Conference on Learning Representations},
year={2025},
url={https://openreview.net/forum?id=MnfHxPP5gs}
}

@inproceedings{stiennon2020summarize,
 author = {Stiennon, Nisan and Ouyang, Long and Wu, Jeffrey and Ziegler, Daniel and Lowe, Ryan and Voss, Chelsea and Radford, Alec and Amodei, Dario and Christiano, Paul F},
 booktitle = {Advances in Neural Information Processing Systems},
 pages = {3008--3021},
 publisher = {Curran Associates, Inc.},
 title = {Learning to summarize with human feedback},
 url = {https://arxiv.org/abs/2009.01325},
 volume = {33},
 year = {2020}
}

@inproceedings{mmlu,
title={{Measuring Massive Multitask Language Understanding}},
author={Dan Hendrycks and Collin Burns and Steven Basart and Andy Zou and Mantas Mazeika and Dawn Song and Jacob Steinhardt},
booktitle={International Conference on Learning Representations},
year={2021},
url={https://openreview.net/forum?id=d7KBjmI3GmQ}
}

@inproceedings{math,
 author = {Hendrycks, Dan and Burns, Collin and Kadavath, Saurav and Arora, Akul and Basart, Steven and Tang, Eric and Song, Dawn and Steinhardt, Jacob},
 booktitle = {Proceedings of the Neural Information Processing Systems Track on Datasets and Benchmarks},
 title = {{Measuring Mathematical Problem Solving With the MATH Dataset}},
 url = {https://openreview.net/forum?id=7Bywt2mQsCe},
 volume = {1},
 year = {2021}
}

@article{sahoo2025llmjudge,
      title={{Quantitative LLM Judges}}, 
      author={Aishwarya Sahoo and Jeevana Kruthi Karnuthala and Tushar Parmanand Budhwani and Pranchal Agarwal and Sankaran Vaidyanathan and Alexa Siu and Franck Dernoncourt and Jennifer Healey and Nedim Lipka and Ryan Rossi and Uttaran Bhattacharya and Branislav Kveton},
      year={2025},
      journal={arXiv preprint arXiv:2506.02945},
      url={https://arxiv.org/abs/2506.02945}, 
}

@inproceedings{vLLM,
author = {Kwon, Woosuk and Li, Zhuohan and Zhuang, Siyuan and Sheng, Ying and Zheng, Lianmin and Yu, Cody Hao and Gonzalez, Joseph and Zhang, Hao and Stoica, Ion},
title = {{Efficient Memory Management for Large Language Model Serving with PagedAttention}},
year = {2023},
isbn = {9798400702297},
publisher = {Association for Computing Machinery},
address = {New York, NY, USA},
doi = {10.1145/3600006.3613165},
booktitle = {Proceedings of the 29th Symposium on Operating Systems Principles},
pages = {611–626},
numpages = {16},
location = {Koblenz, Germany},
series = {SOSP '23}
}

@inproceedings{AdamW,
title={{Decoupled Weight Decay Regularization}},
author={Ilya Loshchilov and Frank Hutter},
booktitle={International Conference on Learning Representations},
year={2019},
url={https://openreview.net/forum?id=Bkg6RiCqY7},
}

@book{koller2009PGM,
  title     = {{Probabilistic Graphical Models: Principles and Techniques}},
  author    = {Daphne Koller and Nir Friedman},
  year      = {2009},
  publisher = {MIT Press},
  series    = {Adaptive Computation and Machine Learning},
  isbn      = {9780262013192},
}

@article{bottou2018survey,
author = {Bottou, L\'{e}on and Curtis, Frank E. and Nocedal, Jorge},
title = {{Optimization Methods for Large-Scale Machine Learning}},
journal = {SIAM Review},
volume = {60},
number = {2},
pages = {223-311},
year = {2018},
doi = {10.1137/16M1080173},
}

@InProceedings{go2023f-div,
  title = 	 {{Aligning Language Models with Preferences through $f$-divergence Minimization}},
  author =       {Go, Dongyoung and Korbak, Tomasz and Kruszewski, Germ\`{a}n and Rozen, Jos and Ryu, Nahyeon and Dymetman, Marc},
  booktitle = 	 {Proceedings of the 40th International Conference on Machine Learning},
  pages = 	 {11546--11583},
  year = 	 {2023},
  volume = 	 {202},
  series = 	 {Proceedings of Machine Learning Research},
  month = 	 {23--29 Jul},
  publisher =    {PMLR},
  url = 	 {https://proceedings.mlr.press/v202/go23a.html},
}

@inproceedings{khalifa2021dm,
title={{A Distributional Approach to Controlled Text Generation}},
author={Muhammad Khalifa and Hady Elsahar and Marc Dymetman},
booktitle={International Conference on Learning Representations},
year={2021},
url={https://openreview.net/forum?id=jWkw45-9AbL}
}

@inproceedings{kim2025guaranteed,
title={{Guaranteed Generation from Large Language Models}},
author={Minbeom Kim and Thibaut Thonet and Jos Rozen and Hwaran Lee and Kyomin Jung and Marc Dymetman},
booktitle={The Thirteenth International Conference on Learning Representations},
year={2025},
url={https://openreview.net/forum?id=8roRgrjbjv}
}

@article{kruszewski2025filtering,
      title={{Whatever Remains Must Be True: Filtering Drives Reasoning in LLMs, Shaping Diversity}}, 
      author={Germán Kruszewski and Pierre Erbacher and Jos Rozen and Marc Dymetman},
      year={2025},
      journal={arXiv preprint arXiv:2512.05962},
      url={https://arxiv.org/abs/2512.05962}, 
}

@Inbook{amari1985,
author="Amari, Shun-ichi",
title={{$\alpha$-Divergence and $\alpha$-Projection in Statistical Manifold}},
bookTitle="Differential-Geometrical Methods in Statistics",
year="1985",
publisher="Springer New York",
address="New York, NY",
pages="66--103",
isbn="978-1-4612-5056-2",
doi="10.1007/978-1-4612-5056-2_3",
}

@article{nielsen2018information,
  title={{What is... An Information Projection?}},
  author={Nielsen, Frank},
  journal={Notices of the AMS},
  volume={65},
  number={3},
  pages={321--324},
  year={2018},
  publisher={American Mathematical Society},
  url={https://www.ams.org/journals/notices/201803/rnoti-p321.pdf?adat=March%202018&trk=201803321&cat=none&type=.pdf}
}

@inproceedings{tldr,
    title = {{TL;DR: Mining Reddit to Learn Automatic Summarization}},
    author = {V{\"o}lske, Michael  and
      Potthast, Martin  and
      Syed, Shahbaz  and
      Stein, Benno},
    booktitle = "Proceedings of the Workshop on New Frontiers in Summarization",
    month = sep,
    year = "2017",
    address = "Copenhagen, Denmark",
    publisher = "Association for Computational Linguistics",
    doi = "10.18653/v1/W17-4508",
    pages = "59--63",
}

\clearpage
\onecolumn
\appendix

\section{Proof of \cref{lem:mukherjee}}
\label{sec:proof}

The lemma follows from simple algebraic manipulations,
\begin{align*}
    \E_{\hat{z}, \hat{y} \sim \pi(\cdot \mid x; \theta)}\!\left[ r(\hat{y}, y^\star) \right] = {} & \E_{\hat{z}, \hat{y} \sim q(\cdot \mid x, y^\star; \theta^{(k - 1)})}\!\left[ r(\hat{y}, y^\star) \frac{\pi(\hat{z}, \hat{y} \mid x; \theta)}{q(\hat{z}, \hat{y} \mid x, y^\star; \theta^{(k - 1)})} \right] \\
    \geq {} & \E_{\hat{z}, \hat{y} \sim q(\cdot \mid x, y^\star; \theta^{(k - 1)})}\!\left[ r(\hat{y}, y^\star) \left(1 + \log \frac{\pi(\hat{z}, \hat{y} \mid x; \theta)}{q(\hat{z}, \hat{y} \mid x, y^\star; \theta^{(k - 1)})} \right) \right] \tag{$z \geq 1 + \log z$} \\
    = {} & \E_{\hat{z}, \hat{y} \sim q(\cdot \mid x, y^\star; \theta^{(k - 1)})}\!\left[ r(\hat{y}, y^\star) \log \pi(\hat{z}, \hat{y} \mid x; \theta) \right] + {} \\
    & \! \underbrace{\E_{\hat{z}, \hat{y} \sim q(\cdot \mid x, y^\star; \theta^{(k - 1)})}\!\left[ r(\hat{y}, y^\star) \left( 1 - \log q(\hat{z}, \hat{y} \mid x, y^\star; \theta^{(k - 1)}) \right) \right]}_{\triangleq C}\,,
\end{align*}
where $C \geq 0$ is independent of $\theta$. The tightness follows from the fact that $z = 1 + \log z$ if and only if $z = 1$. \qed

\newpage

\section{Prompts}
\label{sec:prompts}

The main prompt is shown in \cref{fig:prompt}. The system prompt is:

\begin{tcolorbox}[title=System prompt,halign=left,fontupper=\small,fontlower=\small]
  You are a fair judge assistant tasked with providing clear objective feedback based on specific criteria, ensuring each assessment reflects the absolute standards set for performance.
\end{tcolorbox}

Score rubrics for HelpSteer2 dataset are:

\begin{tcolorbox}[title=Rubric (Helpfulness),halign=left,fontupper=\small,fontlower=\small]
  Score 0: The response is not useful or helpful at all. The response completely missed the essence of what the user wanted. \\
  Score 1: The response is borderline unhelpful and mostly does not capture what the user was looking for, but it is still usable and helpful in a small way. \\
  Score 2: The response is partially helpful but misses the overall goal of the user's query/input in some way. The response did not fully satisfy what the user was looking for. \\
  Score 3: The response is mostly helpful and mainly aligned with what the user was looking for, but there is still some room for improvement. \\
  Score 4: The response is extremely helpful and completely aligned with the spirit of what the prompt was asking for.
\end{tcolorbox}

\begin{tcolorbox}[title=Rubric (Correctness),halign=left,fontupper=\small,fontlower=\small]
  Score 0: The response is completely incorrect. All information provided is wrong, false or hallucinated. If the prompt asks the assistant to do a task, the task is not at all attempted, or the wrong task was attempted in the response. The response is completely irrelevant to the prompt. \\
  Score 1: The response has some correct elements but is mostly wrong or incomplete. The response may contain multiple instances of hallucinations, false information, misleading information, or irrelevant information. If the prompt asks the assistant to do a task, the task was attempted with a small amount of success. \\
  Score 2: The response contains a mix of correct and incorrect information. The response may miss some details, contain misleading information, or minor hallucinations, but is more or less aligned with what the prompt asks for. If the prompt asks the assistant to perform a task, the task is attempted with moderate success but still has clear room for improvement. \\
  Score 3: The response is mostly accurate and correct with a small amount of missing information. It contains no misleading information or hallucinations. If the prompt asks the assistant to perform a task, the task is mostly successfully attempted. \\
  Score 4: The response is completely correct and accurate to what is requested by the prompt with no necessary details missing and without false, misleading, or hallucinated information. If the prompt asks the assistant to do a task, the task is completely done and addressed in the response.
\end{tcolorbox}

\begin{tcolorbox}[title=Rubric (Coherence),halign=left,fontupper=\small,fontlower=\small]
  Score 0: (Completely Incoherent and/or Unclear) – The response is completely incomprehensible and no clear meaning or sensible message can be discerned from it. \\
  Score 1: (Mostly Incoherent and/or Unclear) – The response is mostly hard to follow, with inconsistencies, contradictions, confusing logic flow, or unclear language used throughout, but there are some coherent/clear parts. \\
  Score 2: (A Little Unclear and/or Incoherent) – The response is a little unclear. There are some inconsistencies or contradictions, run on sentences, confusing statements, or hard to follow sections of the response. \\
  Score 3: (Mostly Coherent and Clear) – The response is mostly clear and coherent, but there may be one or two places where the wording is confusing or the flow of the response is a little hard to follow. Over all, the response can mostly be followed with a little room for improvement. \\
  Score 4: (Perfectly Coherent and Clear) – The response is perfectly clear and self-consistent throughout. There are no contradictory assertions or statements, the writing flows logically and following the train of thought/story is not challenging.
\end{tcolorbox}

\begin{tcolorbox}[title=Rubric (Complexity),halign=left,fontupper=\small,fontlower=\small]
  Score 0: (Basic) – The response uses very easy to understand language that is clear and completely interpretable by children, adults, and anyone with a functional command of the language. \\
  Score 1: (Simple) – The response uses relatively straightforward language and wording, but some schooling through elementary or a middle school in the language might be required to understand the response. \\
  Score 2: (Intermediate) – People who have completed up through a high school education will probably be able to understand the vocabulary and sentence structure used, but those at the basic level or children might struggle to understand the response. \\
  Score 3: (Advanced) – The response uses a fairly sophisticated vocabulary and terminology. Someone majoring in this subject at a college or university could have written it and would understand the response. An average adult who does not work or study in this area could not have written the response. \\
  Score 4: (Expert) – An expert in the field or area could have written the response. It uses specific and technically relevant vocabulary. Elevated language that someone at the simple or basic level may not understand at all. The professional language of a lawyer, scientist, engineer, or doctor falls into this category.
\end{tcolorbox}

\begin{tcolorbox}[title=Rubric (Verbosity),halign=left,fontupper=\small,fontlower=\small]
  Score 0: (Succinct) – The response is short, to the point, and the most concise it can be. No additional information is provided outside of what is requested by the prompt (regardless of if the information or response itself is incorrect, hallucinated, or misleading. A response that gives an incorrect answer can still be succinct.). \\
  Score 1: (Pretty Short) – The response is on the shorter side but could still have words, details, and/or text removed before it's at a bare minimum of what the response is trying to convey. \\
  Score 2: (Average Length) – The response isn't especially long or short given what the prompt is asking of the model. The length is adequate for conveying a full response but isn't particularly wordy nor particularly concise. \\
  Score 3: (Moderately Long) – The response is on the longer side but could still have more added to it before it is considered fully detailed or rambling. \\
  Score 4: (Verbose) – The response is particularly lengthy, wordy, and/or extensive with extra details given what the prompt requested from the assistant model. The response can be verbose regardless of if the length is due to repetition and incoherency or if it is due to rich and insightful detail.
\end{tcolorbox}

Score rubrics for Summarize from Feedback dataset are:

\begin{tcolorbox}[title=Rubric (Helpfulness),halign=left,fontupper=\small,fontlower=\small]
  Score 1: The response is terrible. \\
  Score 4: The response is an okay representation of the post, but could be significantly improved. \\
  Score 7: The response is an excellent representation of the post.
\end{tcolorbox}

\begin{tcolorbox}[title=Rubric (Accuracy),halign=left,fontupper=\small,fontlower=\small]
  Score 1: The response is completely wrong, made up, or exactly contradicts what is written in the post. \\
  Score 4: The response says at least one substantial thing that is not mentioned in the post, or that contradicts something in the post. \\
  Score 7: The response has no incorrect statements or misleading implications.
\end{tcolorbox}

\begin{tcolorbox}[title=Rubric (Coverage),halign=left,fontupper=\small,fontlower=\small]
  Score 1: The response contains no information relevant to the post. \\
  Score 4: The response is missing at least 1 important piece of information required to understand the situation. \\
  Score 7: The response covers all of the important information required to understand the situation.
\end{tcolorbox}

\begin{tcolorbox}[title=Rubric (Coherence),halign=left,fontupper=\small,fontlower=\small]
  Score 1: The response is impossible to understand. \\
  Score 4: The response has mistakes or confusing phrasing that make it a bit hard to understand. \\
  Score 7: The response is perfectly clear.
\end{tcolorbox}

\newpage

\section{Ablations Studies}
\label{sec:ablations studies}

\begin{figure*}[t!]
  \centering
  \includegraphics[width=5.6in,bb=0in 8in 7in 10in,clip]{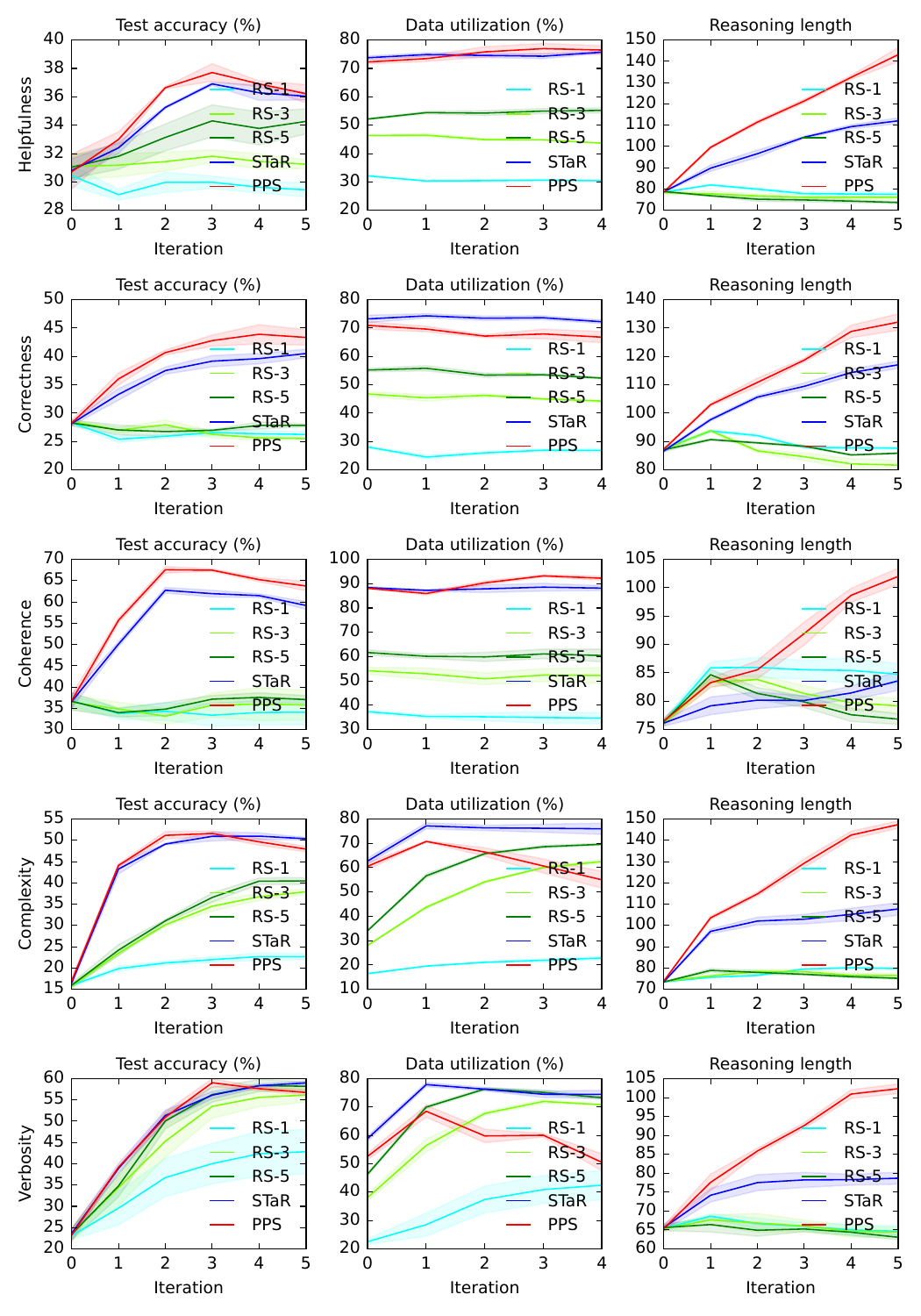}
  \caption{Test accuracy, data utilization, and reasoning length for Qwen 3B and HelpSteer2 helpfulness.}
  \label{fig:rs ablation}
\end{figure*}

We experiment with various rejection sampling budgets $M \in \{1, 3, 5\}$ in \algrs. In \cref{fig:rs ablation}, we show an example where the accuracy clearly increases with $M$. This gain is due to increased data utilization. However, even with this improvement, \algrs is less accurate than both \algpps and \algstar. In addition, its computation time is (in the worst-case) linear in $M$ and thus $M$ times higher than that of \algpps. We summarize the highest accuracies attained by all sampling schemes on HelpSteer2 dataset in \cref{tab:rs ablation}. In all cases, the accuracies of \algpps are higher than those of any rejection sampling scheme.

\begin{table*}[t!]
  \centering
  \small
  \caption{Highest test accuracies for Qwen 3B on all metrics in HelpSteer2 dataset across $M \in \{1, 3, 5\}$ for \algrs.}
  \label{tab:rs ablation}
  \begin{tabular}{l|rrrrr>{\columncolor{highlight}}r} \hline
    Metric & Base & \algrsone & \algrsthree & \algrsfive & \algstar & \algpps \\ \hline
    Helpfulness & 30.44 & 29.97 & 31.81 & 34.30 & 36.91 & \textbf{37.71} \\
    Correctness & 28.30 & 26.59 & 27.96 & 27.86 & 40.51 & \textbf{43.91} \\
    Coherence & 36.75 & 34.28 & 36.00 & 37.56 & 62.73 & \textbf{67.55} \\
    Complexity & 15.94 & 22.68 & 37.91 & 40.43 & 50.98 & \textbf{51.60} \\
    Verbosity & 23.09 & 42.82 & 56.16 & 58.37 & 58.99 & \textbf{59.04} \\ \hline
  \end{tabular}
\end{table*}

\begin{figure*}[t!]
  \centering
  \includegraphics[width=5.6in,bb=0in 4in 7in 6in,clip]{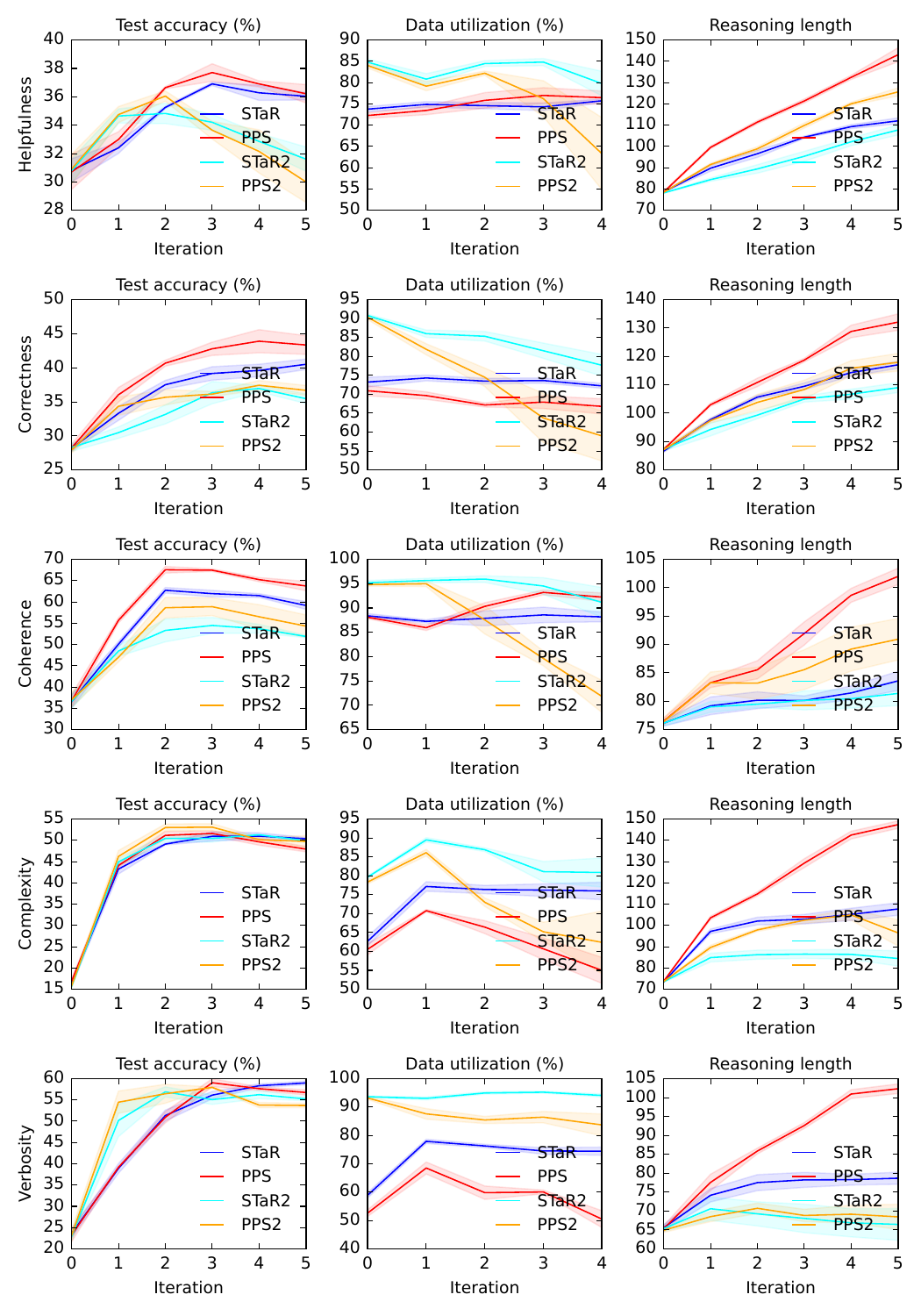}
  \includegraphics[width=5.6in,bb=0in 4in 7in 6in,clip]{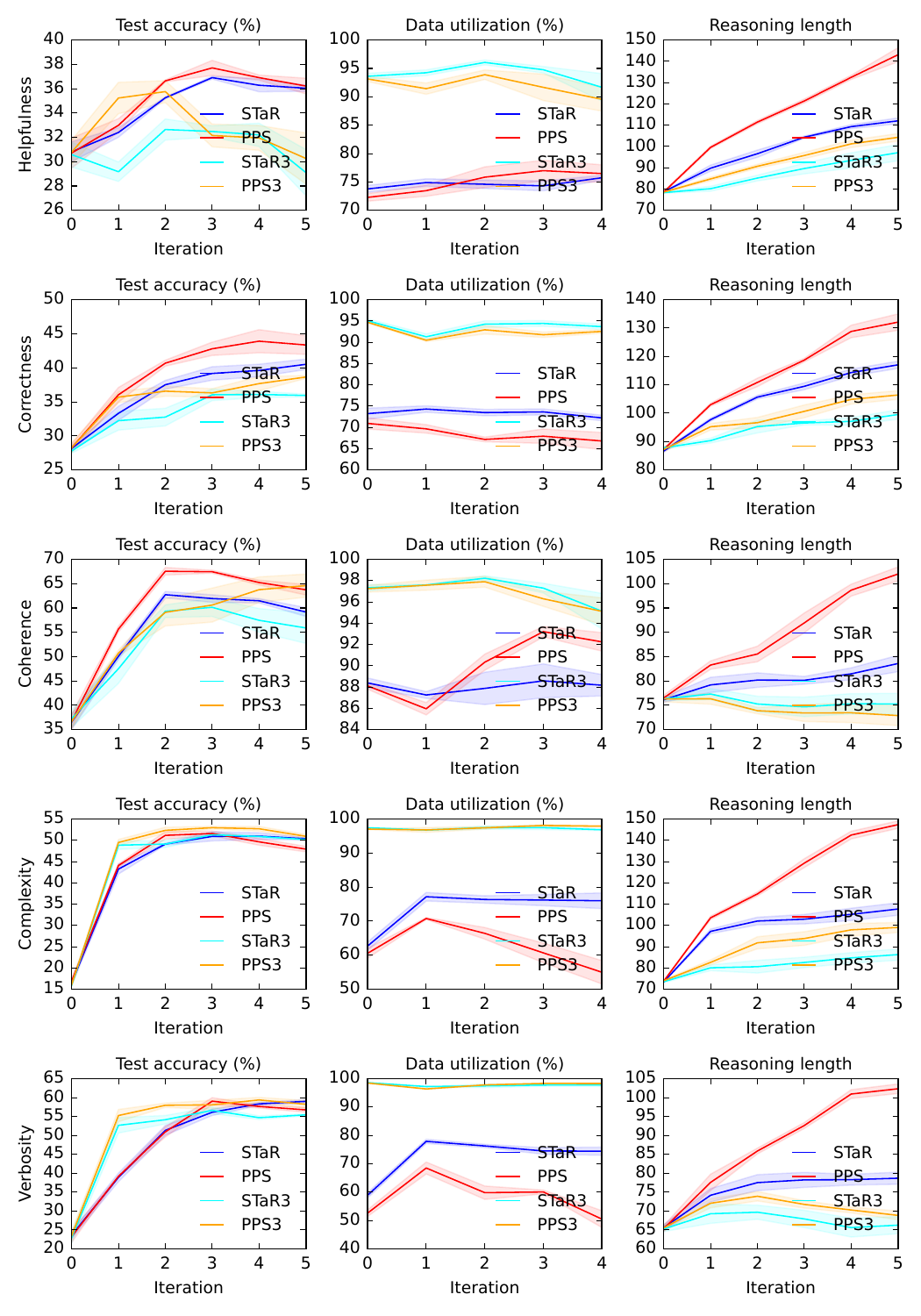}
  \caption{Test accuracy, data utilization, and reasoning length for Qwen 3B and HelpSteer2 coherence.}
  \label{fig:prompt ablation}
\end{figure*}

We further examine robustness to the conditioning prompt. We experiment with two variants of the hint in \cref{fig:prompt}. The first is \say{{\color{cyan}The score of the response should be \textbf{\{conditioning\}}. Do not output the hint.}}, and we denote the corresponding algorithms by \algppstwo and \algstartwo. The second is \say{{\color{cyan}The score of the response should be \textbf{\{conditioning\}}.}}. The corresponding algorithms are denoted by \algppsthree and \algstarthree. We show plots for all prompts and HelpSteer2 coherence in \cref{fig:prompt ablation}. While the absolute accuracy of the new prompts decreases, the relative ranking of the sampling schemes remains unchanged, confirming that our findings generalize beyond a single prompt. Crucially, this accuracy drop highlights that the prompt formulation matters: precise conditioning that explicitly instructs the LLM \emph{how} to use the evidence yields better results than simply stating the evidence as a fact. We summarize the highest accuracies attained by all sampling schemes on HelpSteer2 dataset in \cref{tab:prompt ablation}. In all cases, \algpps has higher accuracies than \algstar, \algppstwo has higher accuracies than \algstartwo, and \algppsthree has higher accuracies than \algstarthree.

\begin{table*}[t!]
  \centering
  \small
  \caption{Highest test accuracies for Qwen 3B on all metrics in HelpSteer2 dataset across different variants of \algpps.}
  \label{tab:prompt ablation}
  \begin{tabular}{l|rrr>{\columncolor{highlight}}rr>{\columncolor{highlight}}rr>{\columncolor{highlight}}r} \hline
    Metric & Base & \algrs & \algstar & \algpps &
    \algstartwo & \algppstwo & \algstarthree & \algppsthree \\ \hline
    Helpfulness & 30.75 & 29.90 & 36.91 & \textbf{37.71} &
    34.83 & 36.05 & 32.64 & 35.76 \\
    Correctness & 28.07 & 26.68 & 40.51 & \textbf{43.91} &
    36.96 & 37.43 & 36.09 & 38.65 \\
    Coherence & 37.07 & 34.74 & 62.73 & \textbf{67.55} &
    54.47 & 58.88 & 60.14 & 64.60 \\
    Complexity & 15.66 & 22.47 & 50.98 & 51.60 &
    51.29 & \textbf{53.09} & 51.29 & 52.96 \\
    Verbosity & 22.91 & 43.13 & 58.99 & 59.04 &
    56.90 & 57.94 & 56.62 & \textbf{59.37} \\ \hline
  \end{tabular}
\end{table*}

\begin{table*}[t!]
  \centering
  \small
  \caption{Highest test accuracies for Qwen 3B on all metrics in HelpSteer2 dataset across different dataset sizes.}
  \label{tab:dataset ablation}
  \begin{tabular}{ll|rrr>{\columncolor{highlight}}r} \hline
    Dataset size $N$ & Metric & Base & \algrs & \algstar & \algpps \\ \hline
    1000 & Helpfulness & 30.74 & 30.87 & 36.84 & \textbf{38.78} \\
    & Correctness & 28.10 & 27.10 & 34.97 & \textbf{37.22} \\
    & Coherence & 36.76 & 35.63 & \textbf{58.89} & 55.43 \\
    & Complexity & 15.66 & 18.79 & \textbf{43.81} & 43.35 \\
    & Verbosity & 22.99 & 30.18 & \textbf{39.94} & 36.68 \\ \hline
    2000 & Helpfulness & 30.38 & 29.99 & 36.91 & \textbf{37.71} \\
    & Correctness & 28.38 & 26.64 & 40.51 & \textbf{43.91} \\
    & Coherence & 37.00 & 34.51 & 62.73 & \textbf{67.55} \\
    & Complexity & 15.99 & 23.12 & 50.98 & \textbf{51.60} \\
    & Verbosity & 23.02 & 42.92 & 58.99 & \textbf{59.04} \\ \hline
    4000 & Helpfulness & 30.42 & 31.79 & \textbf{36.74} & 36.35 \\
    & Correctness & 27.87 & 26.56 & 38.58 & \textbf{40.43} \\
    & Coherence & 36.74 & 39.56 & 62.31 & \textbf{68.83} \\
    & Complexity & 15.66 & 37.05 & 53.04 & \textbf{53.37} \\
    & Verbosity & 23.32 & 56.49 & 59.01 & \textbf{60.34} \\ \hline
  \end{tabular}
\end{table*}

Finally, we experiment with various dataset sizes $N \in \{1\,000, 2\,000, 4\,000\}$ in \cref{tab:dataset ablation}. We observe that when the dataset is much smaller than in our experiment ($N = 1\,000$), both \algstar and \algpps perform poorly and there is no major difference in their accuracies. On the other hand, when the dataset is much larger ($N = 4\,000$), both \algstar and \algpps perform comparably to $N = 2\,000$ and our earlier reported trends hold.

\newpage

\section{Detailed \fem Runs}
\label{sec:detailed fem runs}

\begin{figure*}[!h]
  \centering
  \includegraphics[width=5.6in]{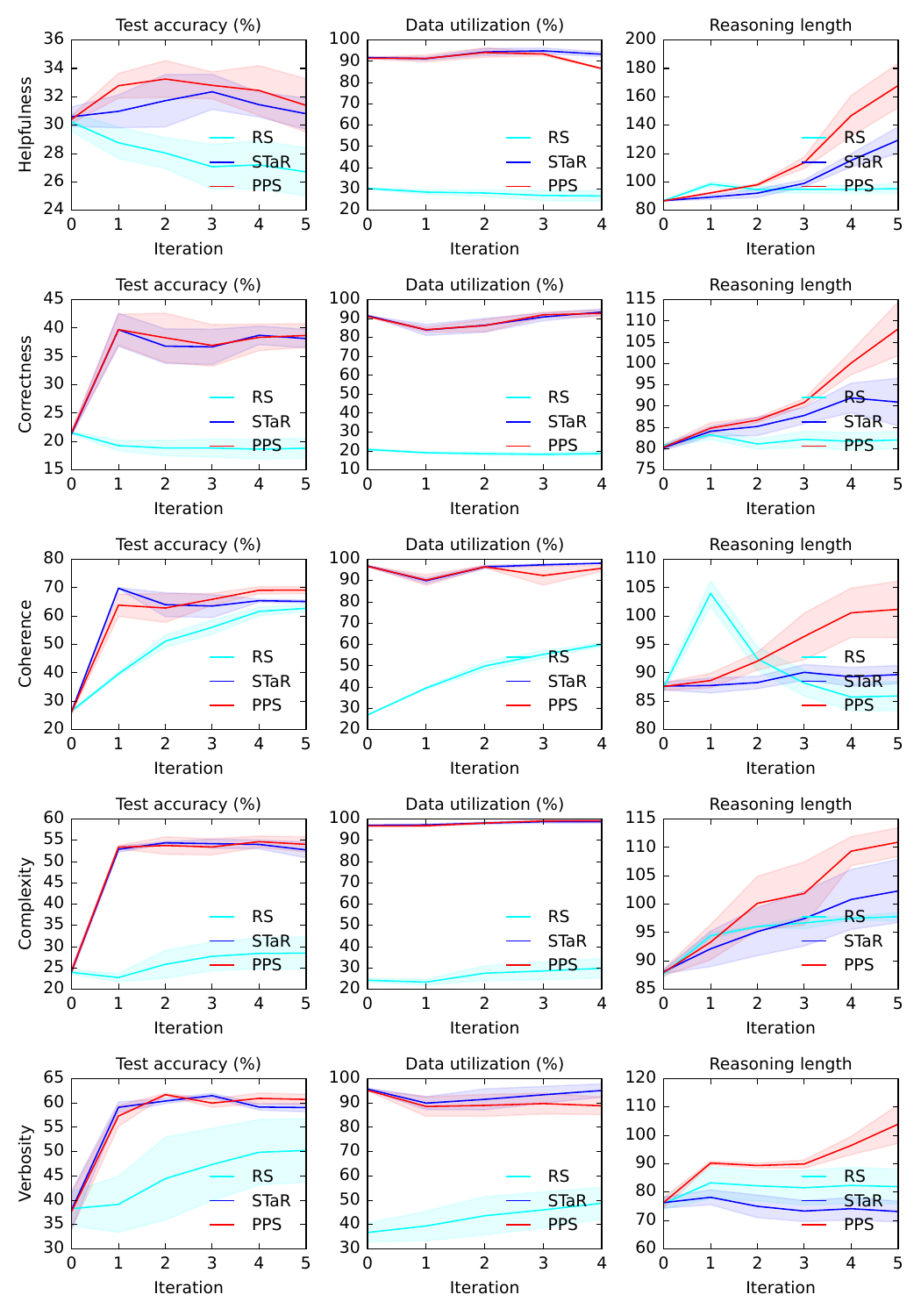}
  \caption{Test accuracy, data utilization, and reasoning length for Llama 3B and HelpSteer2 dataset.}
  \label{fig:hs2 l3b}
\end{figure*}

\begin{figure*}[t!]
  \centering
  \includegraphics[width=5.6in]{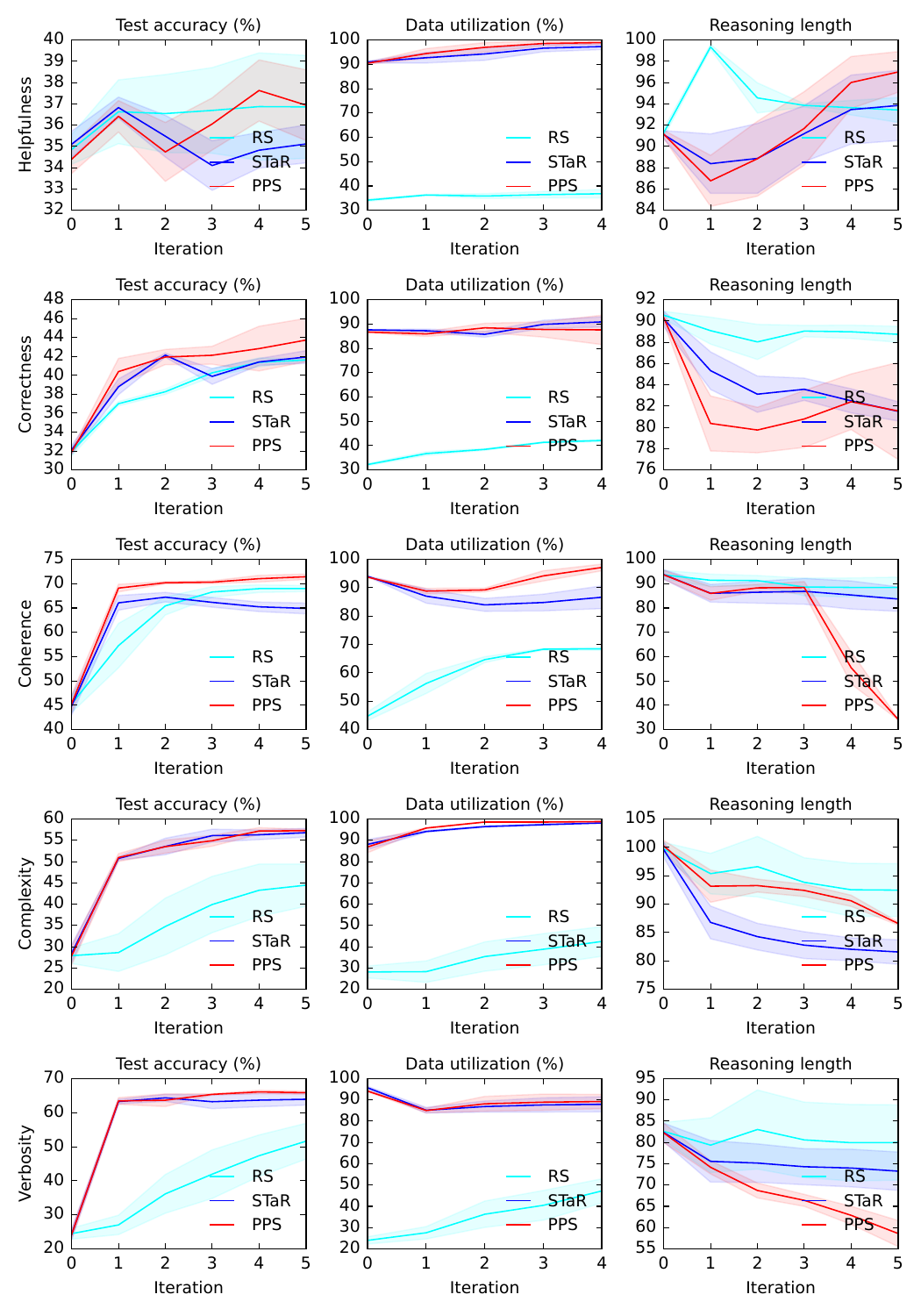}
  \caption{Test accuracy, data utilization, and reasoning length for Llama 8B and HelpSteer2 dataset.}
  \label{fig:hs2 l7b}
\end{figure*}

\begin{figure*}[t!]
  \centering
  \includegraphics[width=5.6in]{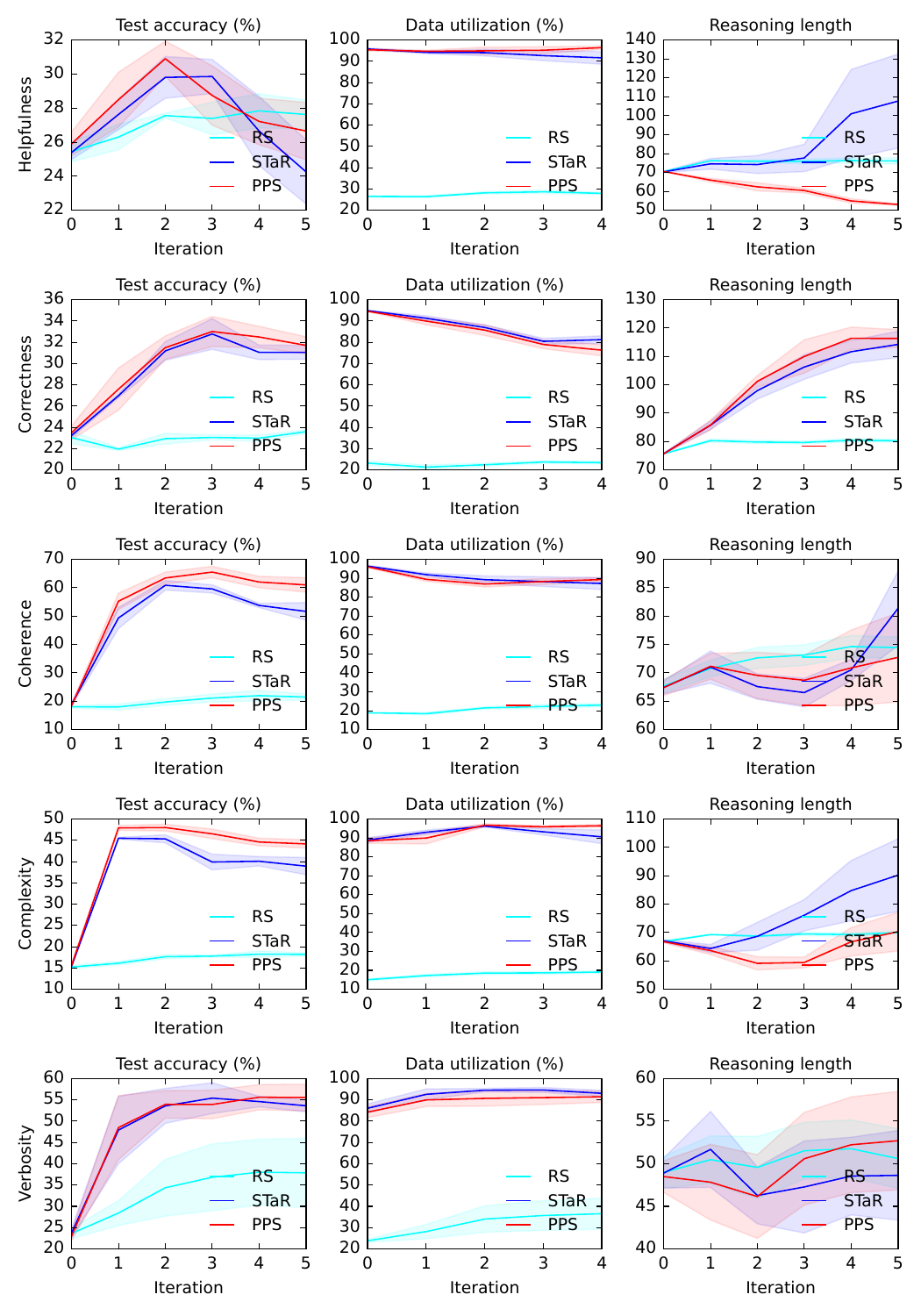}
  \caption{Test accuracy, data utilization, and reasoning length for Qwen 1B and HelpSteer2 dataset.}
  \label{fig:hs2 q1b}
\end{figure*}

\begin{figure*}[t!]
  \centering
  \includegraphics[width=5.6in]{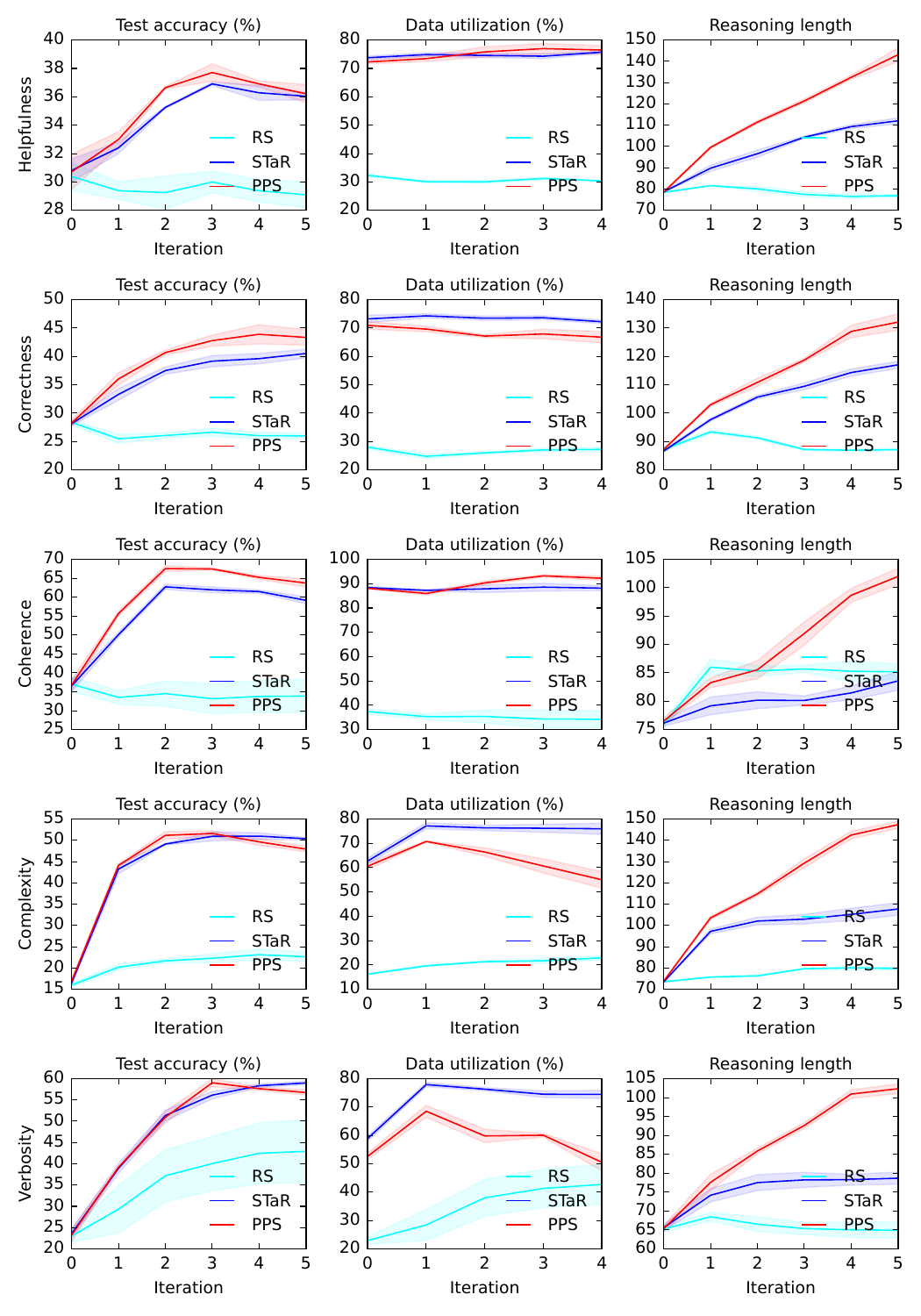}
  \caption{Test accuracy, data utilization, and reasoning length for Qwen 3B and HelpSteer2 dataset.}
  \label{fig:hs2 q3b}
\end{figure*}

\begin{figure*}[t!]
  \centering
  \includegraphics[width=5.6in]{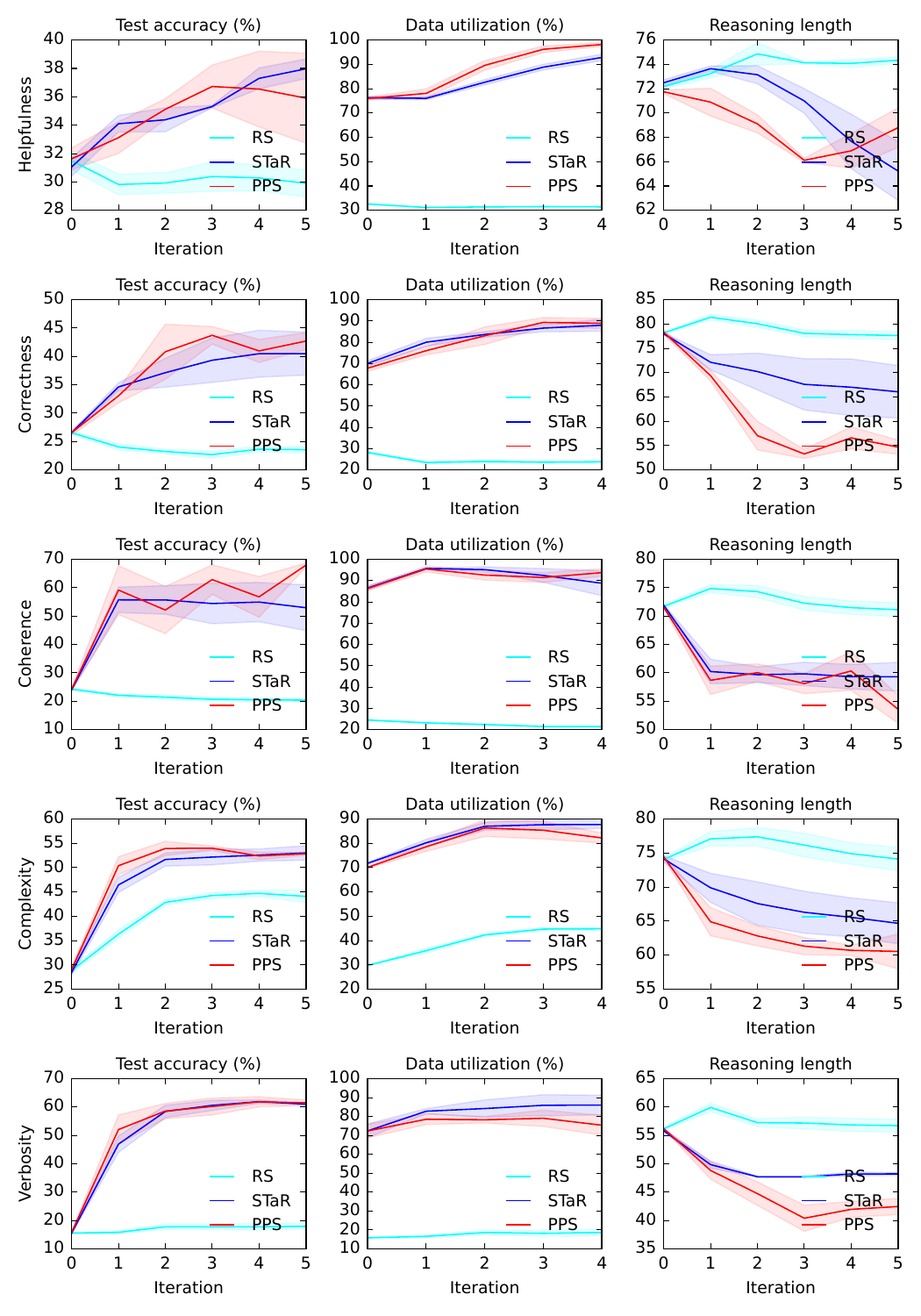}
  \caption{Test accuracy, data utilization, and reasoning length for Qwen 7B and HelpSteer2 dataset.}
  \label{fig:hs2 q7b}
\end{figure*}

\begin{figure*}[t!]
  \centering
  \includegraphics[width=5.6in]{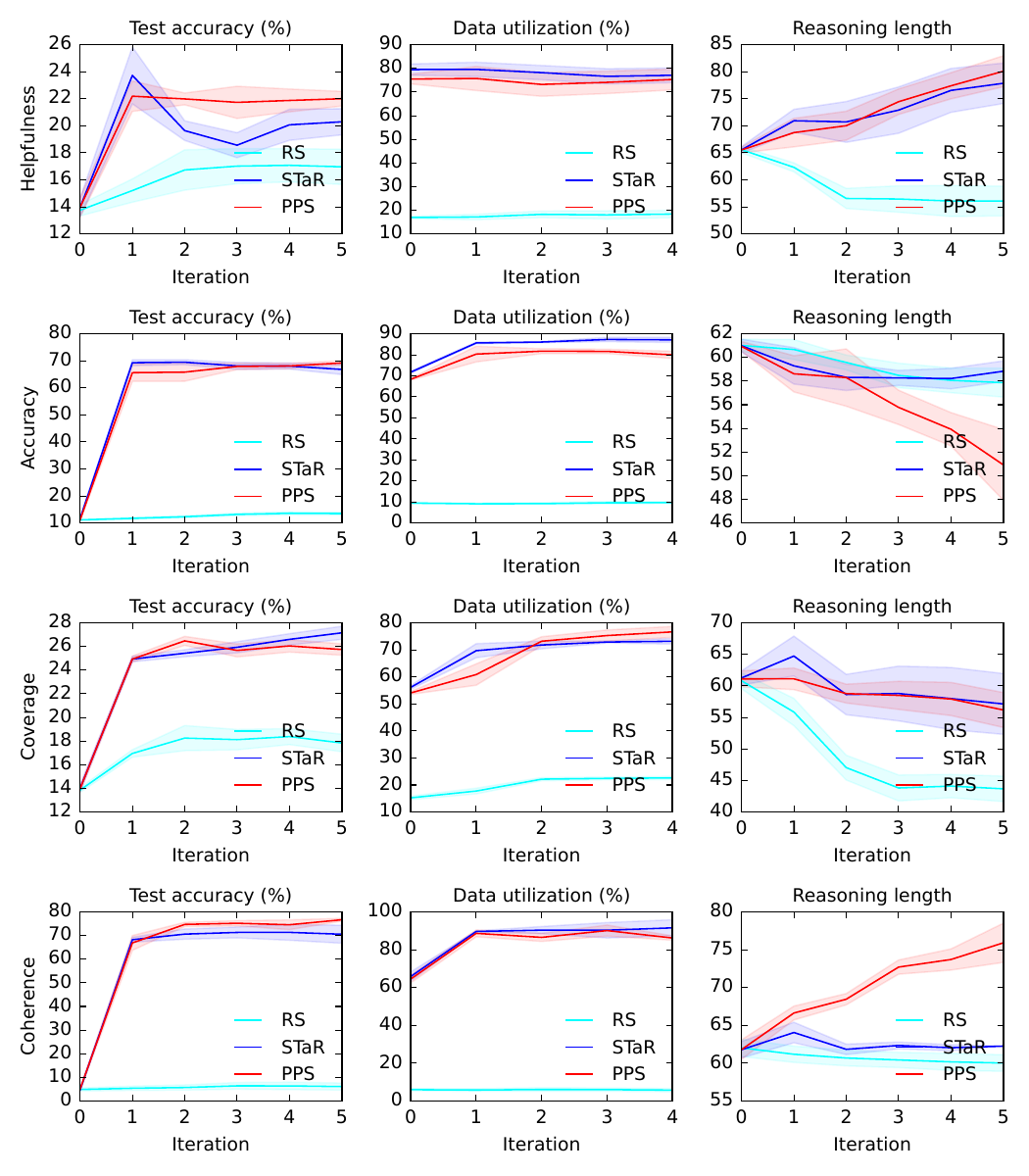}
  \caption{Test accuracy, data utilization, and reasoning length for Llama 3B and Summarize from Feedback dataset.}
  \label{fig:sff l3b}
\end{figure*}

\begin{figure*}[t!]
  \centering
  \includegraphics[width=5.6in]{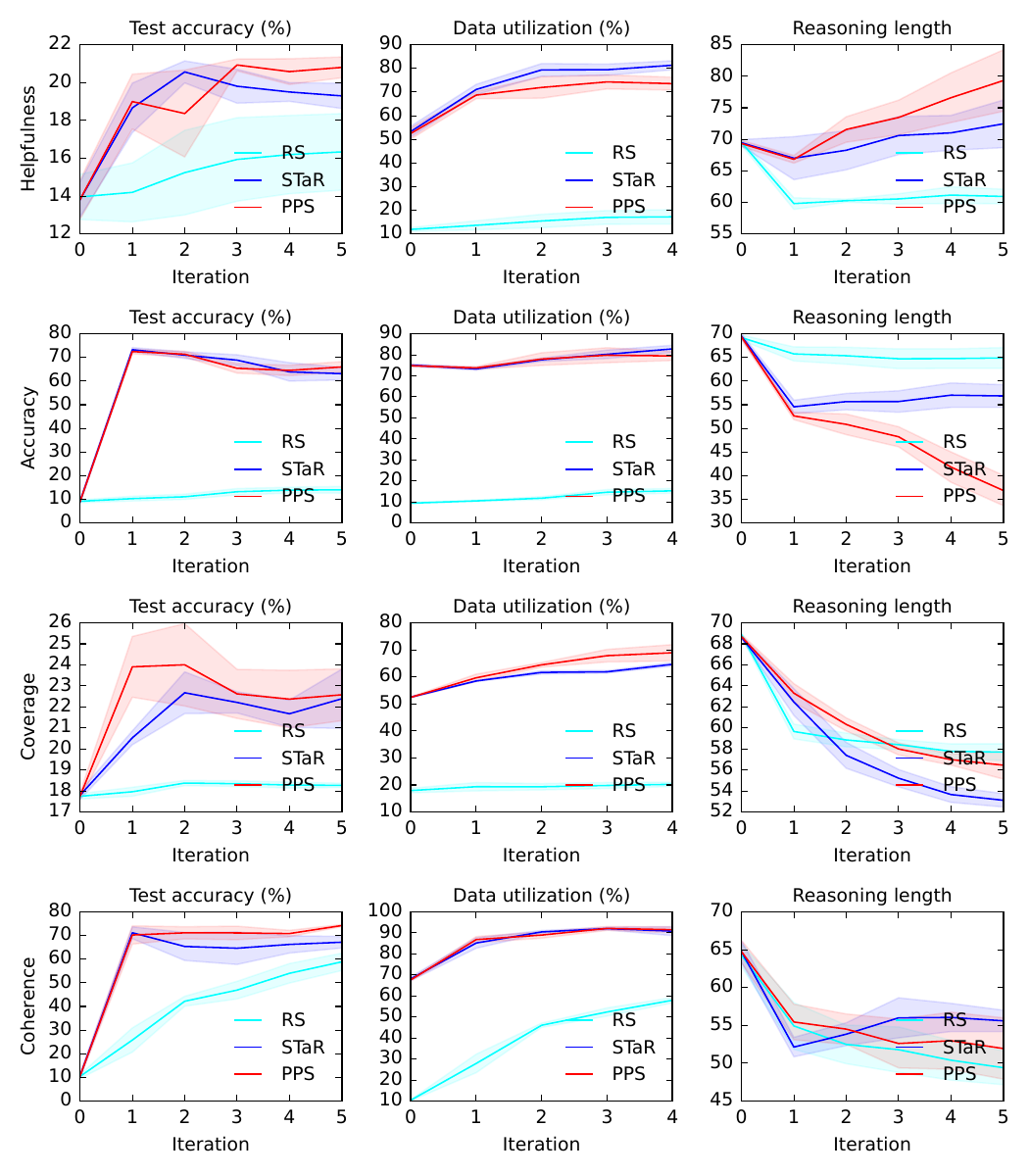}
  \caption{Test accuracy, data utilization, and reasoning length for Llama bB and Summarize from Feedback dataset.}
  \label{fig:sff l7b}
\end{figure*}

\begin{figure*}[t!]
  \centering
  \includegraphics[width=5.6in]{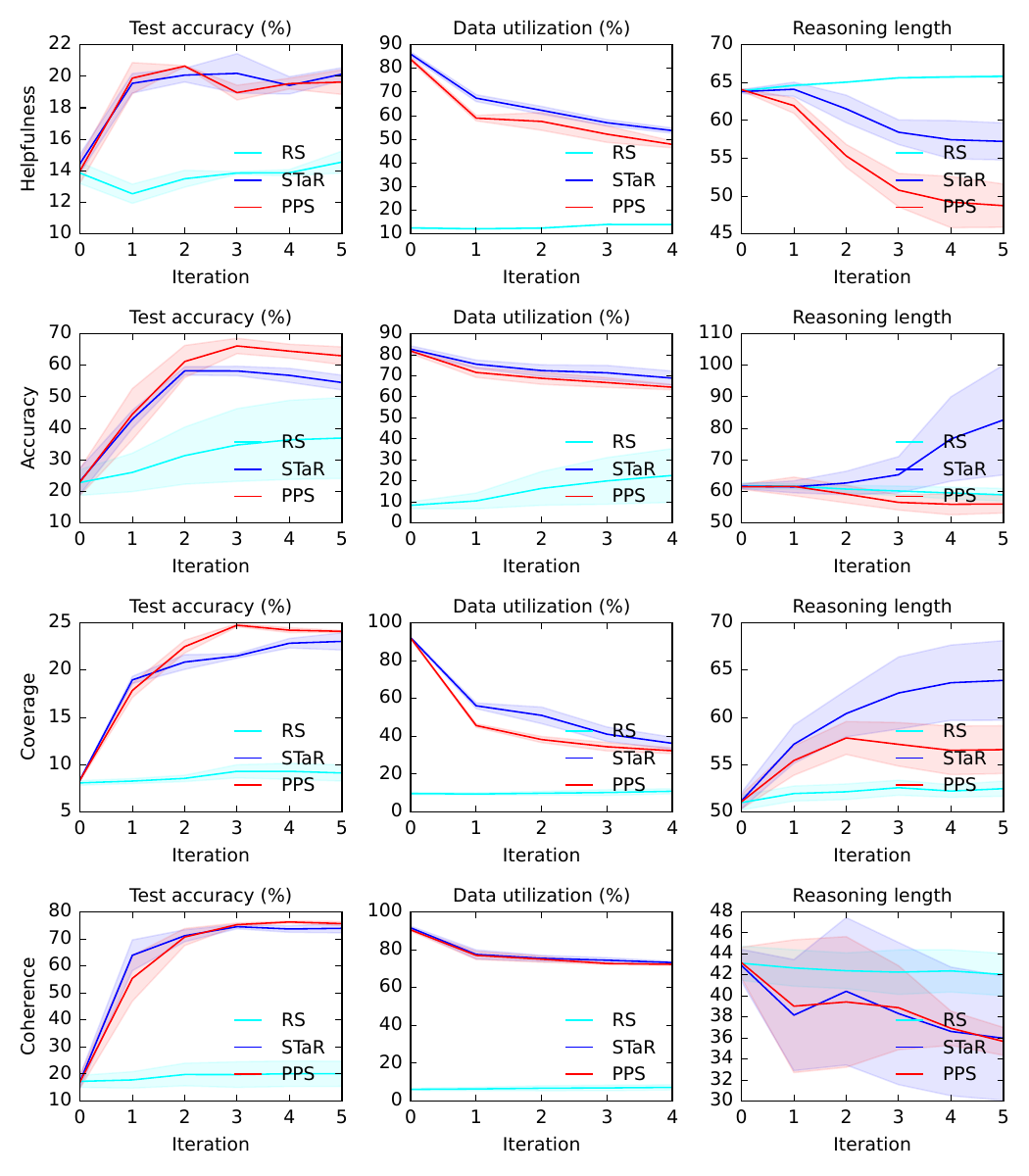}
  \caption{Test accuracy, data utilization, and reasoning length for Qwen 1B and Summarize from Feedback dataset.}
  \label{fig:sff q1b}
\end{figure*}

\begin{figure*}[t!]
  \centering
  \includegraphics[width=5.6in]{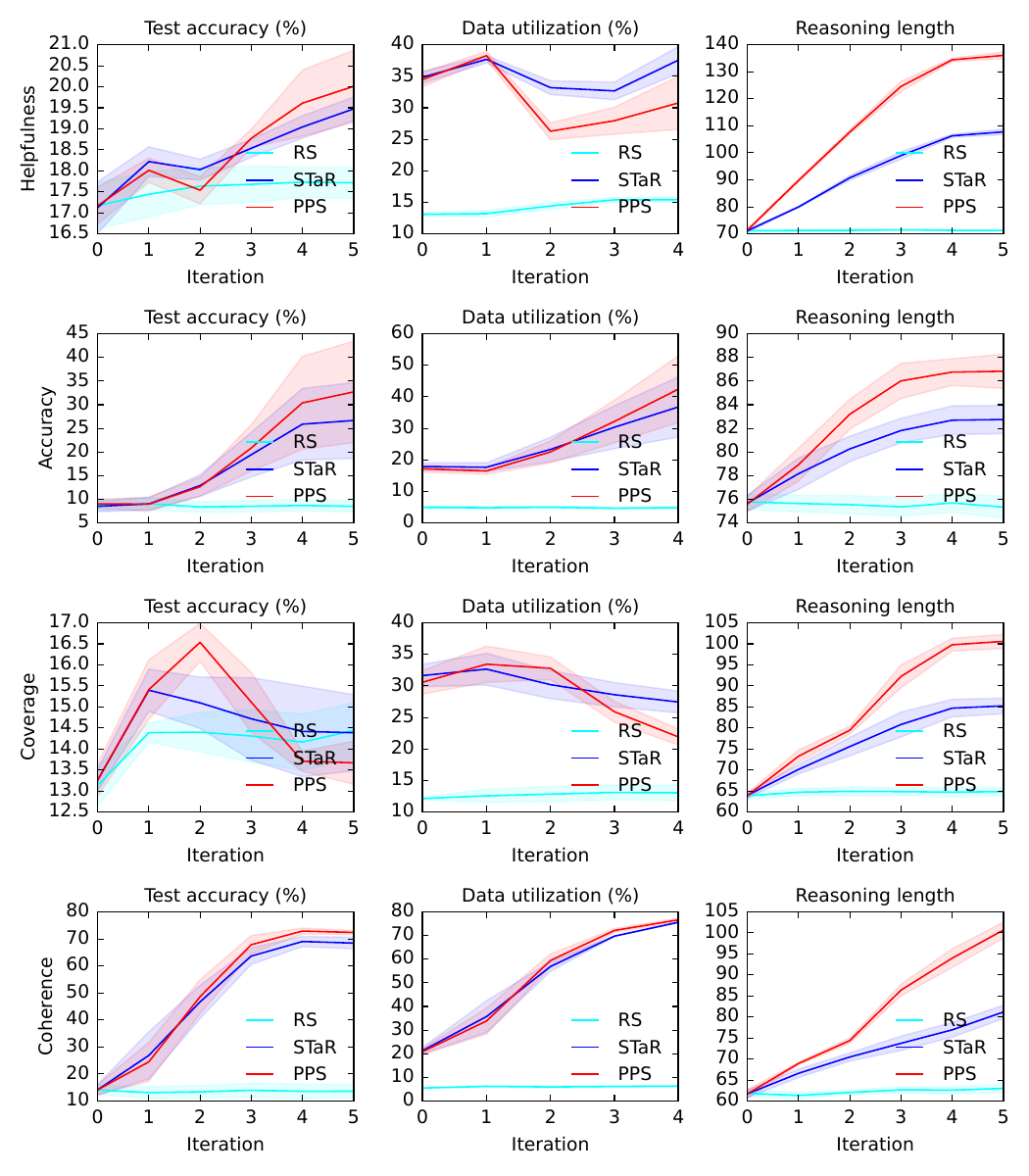}
  \caption{Test accuracy, data utilization, and reasoning length for Qwen 3B and Summarize from Feedback dataset.}
  \label{fig:sff q3b}
\end{figure*}

\begin{figure*}[t!]
  \centering
  \includegraphics[width=5.6in]{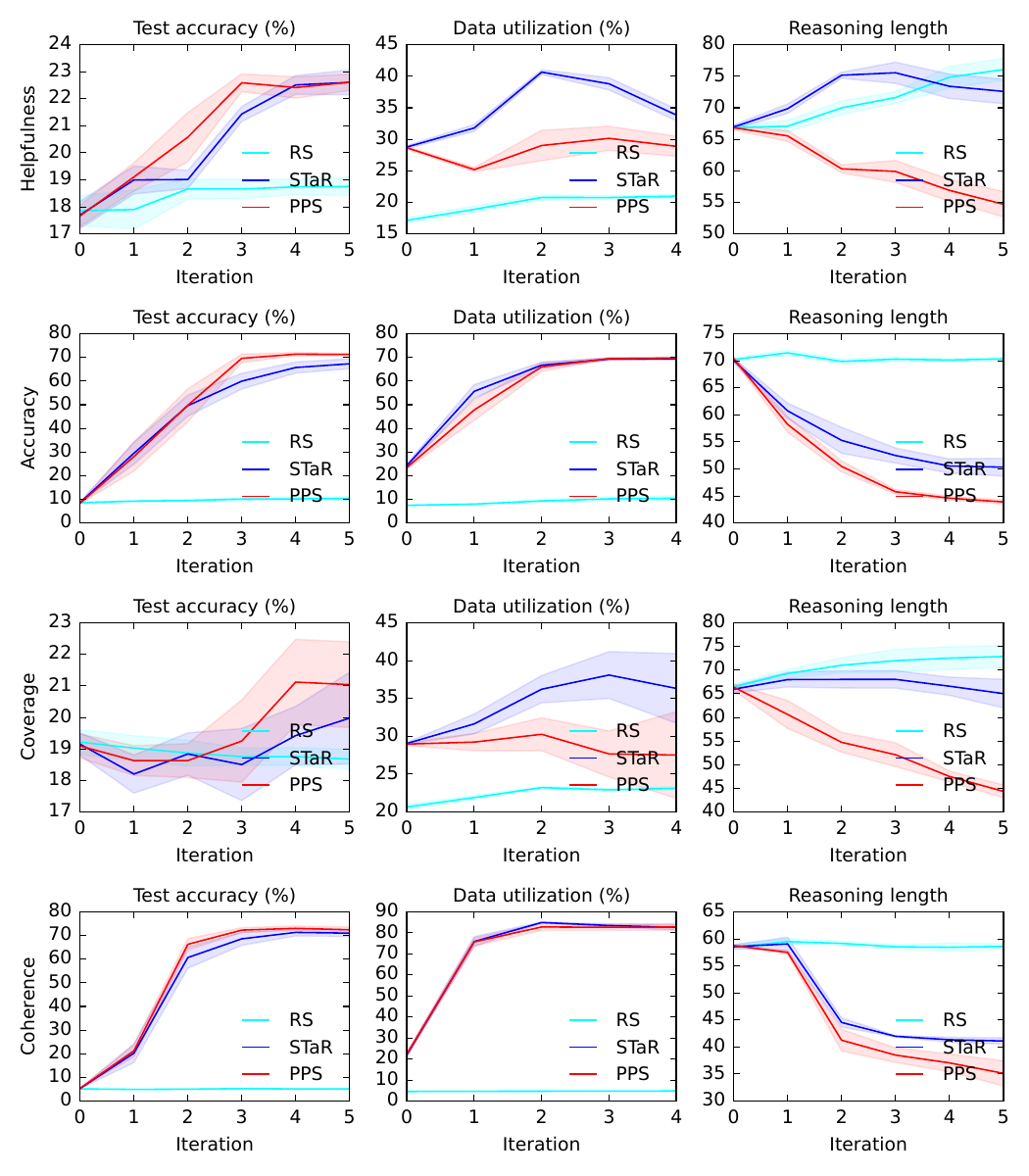}
  \caption{Test accuracy, data utilization, and reasoning length for Qwen 7B and Summarize from Feedback dataset.}
  \label{fig:sff q7b}
\end{figure*}

\end{document}